\pdfoutput=1

\documentclass[11pt]{article}

\usepackage[final]{acl}

\usepackage{times}
\usepackage{latexsym}

\usepackage[T1]{fontenc}

\usepackage[utf8]{inputenc}

\usepackage{microtype}

\usepackage{inconsolata}

\usepackage{graphicx}

\usepackage{amsmath}
\usepackage{amssymb}
\usepackage{amsfonts}
\usepackage{amsthm}
\usepackage{algorithm}
\usepackage{algpseudocode}
\usepackage{hyperref}
\usepackage{tabularx}
\usepackage{tcolorbox, soul} 
\usepackage{listings}
\lstdefinestyle{Bash}
{language=bash,
keywordstyle=\color{black},
basicstyle=\ttfamily,
stringstyle=\color{black},
showstringspaces=false
}
\usepackage{booktabs} 
\usepackage[capitalize,noabbrev]{cleveref}

\newcommand{\methodfull}{Storytelling With Action Guidance}
\newcommand{\methodshort}{\textsc{SWAG}}

\title{\methodshort{}: \methodfull}

\author{Jonathan Pei* 
  \quad
  Karim El-Refai* 
  \quad
  Zeeshan Patel* 
  \quad
  Tianle Li 
  \\
  UC Berkeley
  \\
  \normalfont{\texttt{\{jonnypei, karim.el-refai, zeeshanp, tianleli\}@berkeley.edu}}
  }

\begin{document}
\maketitle

\def\thefootnote{*}\footnotetext{Equal Contribution}\def\thefootnote{\arabic{footnote}}

\begin{abstract}

Automated long-form story generation typically employs long-context large language models (LLMs) for one-shot creation, which can produce cohesive but not necessarily engaging content. We introduce \methodfull{} (\methodshort{}), a novel approach to storytelling with LLMs. Our approach frames story writing as a search problem through a two-model feedback loop: one LLM generates story content, and another auxiliary LLM is used to choose the next best ``action'' to steer the story's future direction. Our results show that \methodshort{} can substantially outperform previous end-to-end story generation techniques when evaluated by GPT-4 and through human evaluation. Our \methodshort{} pipeline using only small open-source models surpasses GPT-3.5-Turbo.

\end{abstract}

\section{Introduction}

Large language models (LLMs) have recently changed the landscape of content generation. A number of works have proposed techniques for short story generation \cite{fan2018hierarchical, wilmot2021temporal, rashkin2020plotmachines, xu2018skeletonbased}. However, it has been a major challenge for AI to generate long-form stories that are both \emph{coherent} and \emph{interesting} \cite{oatley-1995-book,charniak2004storycomp,alabdulkarim2021automatic}. This remains a challenge with SoTA LLMs such as GPT-4 \cite{OpenAI_GPT4_2023}, Llama-2 \cite{touvron2023llama}, and Mistral \cite{jiang2023mistral}.


We propose \methodshort{}, an algorithm for iteratively generating engaging and captivating stories using LLMs. In our work, we structure storytelling as a search problem. This paradigm allows us to formulate the problem as finding the ``optimal path'' in a search space of possible stories given a story idea. By having another model guide the LLM during the story writing process, we can improve control over the story direction and create more engaging content. At a high level, we train an \textit{action discriminator} LLM (AD LLM) to determine the next best action to take given the current state of a story. We then prompt another LLM to write the next part of the story based on the chosen action. This feedback loop generates long-form stories that are fascinating and amusing to read. The main component of our system is the AD LLM, which helps pave the path for the story by selecting the next best ``action'' to continue the story. The AD LLM can be paired with any open-source model (e.g. Llama-2-7B) or closed models (e.g. OpenAI's GPT-4) for generating the story. Our algorithm allows fine-grain control over story content progression while providing the flexibility to integrate custom models for writing the story or using LLM services offered by other companies through APIs.

All code is available at \texttt{\url{https://github.com/jonnypei/swag-storytelling}}.

\section{Related Work}

Prior works have attempted to improve the quality and/or diversity of story generations in a variety of ways.

\subsection{Storytelling with reinforcement learning}

In the context of content generation, reinforcement is largely used for fine-tuning \cite{chang2023learning, bai2022training} or auxiliary model guidance \cite{peng2022guiding, castricato2022robust}.

Perhaps most similar to our work are methods that involve dynamic inference-time option-selection and/or classification \cite{alabdulkarim2021goaldirected, Tambwekar_2019, peng2022guiding}. Our approach differs from prior ones in that our model (1) uses an adapted LLM to interpret an internal representation of the current story; (2) is highly modular; and (3) is prompting-based. These aspects contribute to our method's diverse story generations despite having such a simple, flexible structure.

\subsection{Controlled Text Generation (via prompting)}

The recent advancements in language models have substantially increased the popularity of (simpler) prompting approaches such as chain of thought. Prompts may be manually designed~\cite{gpt3} or automatically designed~\cite{shin2020autoprompt,zou-2021-inverse}; prompting may also be an iterative process~\cite{wei-chainofthought}. 
Some works such as ~\cite{qin2021learning,lester2021power} also explore continuous soft prompts. 
Compared to prior work, our contribution is an iterative feedback-prompting-based method that utilizes an auxiliary LLM for control, enabling more diverse storytelling. 

\subsection{Human-in-the-loop story generation}

As opposed to automatic story generation, some previous works use human-in-the-loop methods to generate interesting long stories \cite{goldfarb-tarrant-etal-2019-plan,coenen2021wordcraft,chung2022talebrush, mirowski2022cowriting,martin2017improvisational,wang2023playing, Lin2021PlugandBlendAF}. We emphasize that although our method is completely automatic without any human intervention, the flexibility of the AD's action space makes it quite intuitive for a human collaborator to ``tune'' our method towards their own liking.




\section{Methods}

\begin{figure}[!ht]
    \centering
    \includegraphics[width=\columnwidth,trim={2cm 10cm 2cm 8cm},clip]{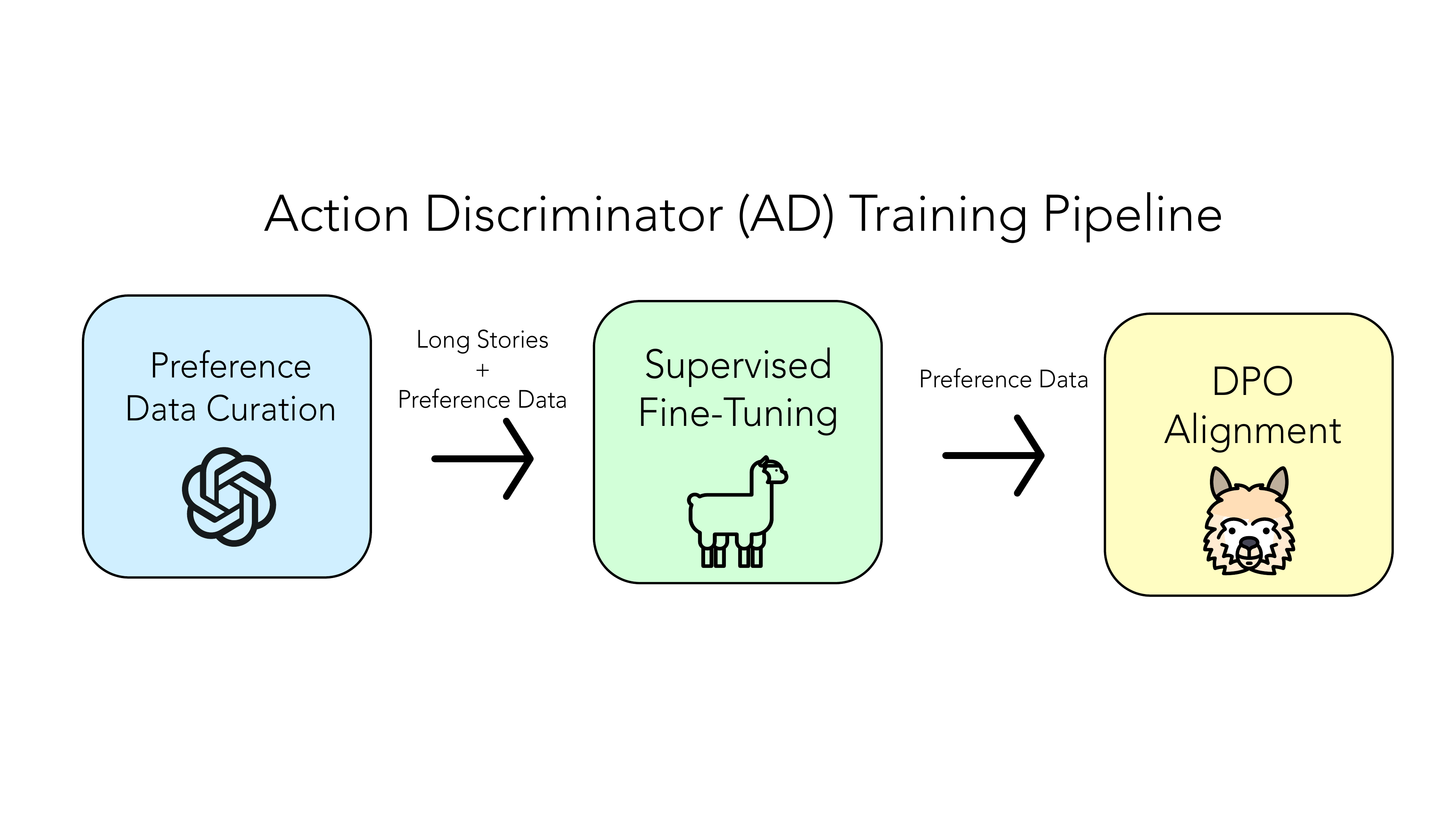}\caption{\textbf{\methodshort{} AD LLM Training Pipeline.} After curating long story and action preference data from GPT-4, we perform SFT on a base open-source LLM, and then align our model with more preference data using DPO to produce our action discriminator model (AD LLM).}
    \label{fig:swagtraining}
\end{figure}

Our creative storytelling method consists of two primary components: the story generation model and the action discriminator model (AD LLM). \methodshort{} enables the use of any open-source LLM or LLM service for story generation. We create an AD LLM by collecting preference data for story actions, and aligning a pretrained LLM on our preference dataset. We visualize our training pipeline in Figure \ref{fig:swagtraining}.

\subsection{Preference Data Collection}

We use a preference dataset of story actions to train a model to learn how to choose an action for the next part of the story. Given a list of actions, we want our AD LLM to select the best action that will keep the reader engaged with the story. Several datasets contain thousands of story prompts and ideas, but there are no preference datasets for choosing the next direction for a story.

To generate this data efficiently, we developed a pipeline that prompted OpenAI's GPT-4 to choose the next best action given a ``story state''. We define the story state to be $$X = (\mathcal{P}, \mathcal{S}, \mathcal{A}),$$ where $\mathcal{P}$ is the story prompt, $\mathcal{S}$ is the current continuation of the story prompt, and $
\mathcal{A}$ is the next ``action'' to take for developing the next part of the story. 

To generate our dataset, we first extract a random subset of the Writing Prompts \cite{fan2018hierarchical} dataset to acquire a diverse set of story prompts. Then, for each story prompt from this subset, we feed it into GPT-4 to write an initial paragraph $S$, forming the dataset 
\[\mathcal{D} = \left\{\left(P_i, S_i, \emptyset\right)\right\}_{i=1}^n\]These story states provide a simple yet comprehensive starting point for the AD LLM to find the best path to continue generating the given story. Note that we use GPT-4 to generate the initial paragraphs because it is one of the most capable LLMs available.

After constructing the initial story states in $\mathcal{D}$, we generate preference data on the next best action for continuing the story. We model this preference data by having a ``chosen'' and ``rejected'' action for each story state. For any given story state $\mathcal{S}$, the chosen action $c$ is what we would like the LLM to choose when deciding the next best direction for the story, and the rejected action $r$ is the path we would like the LLM to avoid for the next part of the story. This preference data allows our model to understand how to rank different actions for the diverse set of story prompts that it will encounter during test-time.

To generate the ranking data, we prompt GPT-4 with an initial story state $\mathcal{S}$ and a list of ``actions'' $\mathcal{A}$ to choose the best direction for the next paragraph in the story. The action used by GPT-4 to generate the next paragraph is set as the chosen action, and we then randomly choose an action from the remaining actions as the rejected action. We distill multiple datasets for supervised fine-tuning (SFT), direct preference optimization (DPO), and evaluation.

\subsection{Supervised Fine-Tuning (SFT)}
In the SFT phase, we follow the typical set up of starting with a pre-trained LLM and fine-tuning it with supervised learning, effectively using a maximum likelihood objective. We fine-tune the LLM on our downstream task of action discrimination on the preference dataset we created using GPT-4. 

We conduct SFT in two stages. During the first stage, we fine-tune the AD LLM on a dataset of long stories. We train the model to take a prompt as an input and generate a long-context story. This process ensures that models like Llama-2-7B, with their shorter default context length, can accurately process longer data sequences. In the second stage, we fine-tune our new long-context AD LLM on a preference dataset with chosen and rejected actions for the next story direction. This stage helps the model better understand the downstream task for which we want to build a preference model. In order to process longer stories, we extend the context length using the technique from LongLoRA \cite{chen2023longlora}.


\subsection{Direct Preference Optimization (DPO)}
We utilize DPO to further refine the results of our action discriminator model. In DPO, we want our policy $\pi_{\text{SFT}}$ to learn how to rank chosen responses $c^{(k)}$ over rejected responses $r^{(k)}$ in a preference model framework. In PPO, we use a learned reward model $R_{\theta}(x, y)$ for which we estimate parameters by taking the maximum likelihood over our static preferences dataset. DPO instead allows us to define a mapping from the optimal reward model to our language model policy, enabling the training of our language model to satisfy our preferences directly with a single cross-entropy loss \cite{rafailov2023direct}. Using DPO, we can refine the SFT model on our preferences dataset to generate actions that are better aligned with the actions chosen by GPT-4.




\subsection{\methodshort{} Feedback Loop}
The main algorithm in our method is the \methodshort{} feedback loop that enables the action guidance mechanism. This feedback loop is a three step process and can be configured to use open-source LLMs, closed-source LLMs, or a hybrid of both for inference (beyond story generation). 

First, we generate an initial story state $X^{(0)} = (\mathcal{P}, \mathcal{S}^{(0)}, \emptyset)$ by passing the story prompt $\mathcal{P}$ into the story generation model $\pi_{\text{story}}$ to yield 
the initial paragraph $\mathcal{S}^{(0)}$. Next, we pass $X^{(0)}$ into our AD LLM $\pi_{\text{AD}}$ along with a list of (predefined) possible actions (included in Appendix ~\ref{app:actions}), and $\pi_{\text{AD}}$ generates the next best action to continue the story. 

After generating the next best action, we update our story state to be \[X^{(0)} = (\mathcal{P}, \mathcal{S}^{(0)}, \mathcal{A}^{(0)}).\]
To generate the story, we iteratively repeat this process of (1) generating the next paragraph in the story via $\pi_{\text{story}}$ and (2) generating the optimal subsequent action to take via $\pi_{\text{AD}}$.
See Algorithm \ref{alg:swag} for a pseudocode implementation of the \methodshort{} feedback loop.
\vspace{-0.5em}

\begin{algorithm}[H]
\caption{\label{alg:swag}Storytelling With Action Guidance (\methodshort{})}
\begin{algorithmic}
\Procedure{\methodshort{}}{$\mathcal{P}, \pi_{\text{story}}, \pi_{\text{AD}}, k$}
\State $\mathcal{S}^{(0)} \gets \pi_{\text{story}}(\mathcal{P})$
\State $\mathcal{A}^{(0)} \gets \pi_{\text{AD}}(\mathcal{P}, \mathcal{S}^{(0)})$
\State $X^{(0)} \gets (\mathcal{P}, \mathcal{S}^{(0)}, \mathcal{A}^{(0)})$
\For{$i = 1 \dots k$}
    \State $\mathcal{S}^{(i)} \gets \mathcal{S}^{(i-1)} + \pi_{\text{story}}(X^{(i-1)})$
    \State $\mathcal{A}^{(i)} \gets \pi_{\text{AD}}(\mathcal{P}, \mathcal{S}^{(i)})$
    \State $X^{(i)} \gets (\mathcal{P}, \mathcal{S}^{(i)}, \mathcal{A}^{(i)})$
\EndFor
\State \Return {$\mathcal{S}^{(k)}$}
\EndProcedure
\end{algorithmic}
\end{algorithm}
\vspace{-1em}

The \methodshort{} feedback loop can be ran as many times as needed until the desired story length is reached—we can freely choose $k$. This feedback mechanism can be implemented between any two LLMs (for story and AD), allowing for enhanced modularity in content generation for stories.

\section{Experiments}

\subsection{Experimental Setup}
In our experiments, we aimed to evaluate the quality of the stories generated by our inference pipeline with different combinations of models and AD settings. We also explored if GPT-4 had any bias in ranking story actions for the preference dataset and the effects of this bias on our AD LLM. 

\subsection{Dataset}
In order to train an AD LLM that can process long-form content, we fine-tuned our model on a dataset of long stories. We distilled this dataset of long stories from Llama-2-7B, Mistral-7B, and Mixtral-8$\times$7B using a sample of prompts from the WritingPrompts dataset. We generated 20,000 long stories from these models, providing a diverse distribution of stories for SFT. We fine-tuned Llama-2-7B and Mistral-7B on this long stories dataset, allowing them to have a context length of 32,768 tokens. 

\begin{figure}[!ht]
    \centering
    \includegraphics[width=3in]{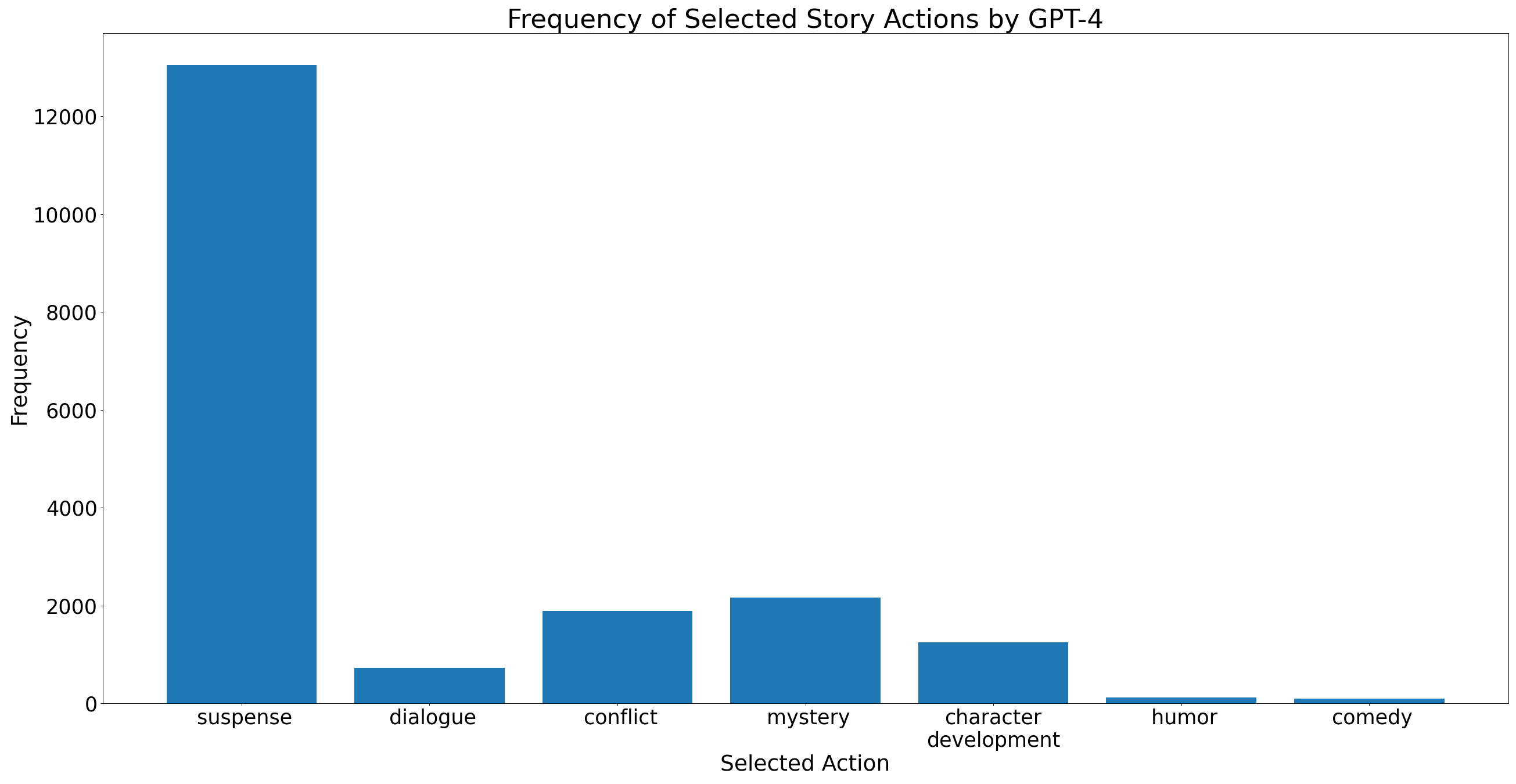}
    \caption{\textbf{Original Distribution of Actions.} We observe a severe distribution imbalance where the vast majority of actions selected is ``add suspense''. \textit{Note: actions chosen with frequency less than 100 not shown.}}
    \label{fig:ogacfreq}
\end{figure}

For our DPO preference dataset, we prompted GPT-4 to generate preference data on a sample of approximately 60,000 prompts from the WritingPrompts dataset. One key aspect of this preference data is the potential options for story actions. We distilled a list of 50 different story actions from GPT-4 and used this set of actions for all training experiments. Some examples of actions in the set include ``add suspense'', ``add mystery'', ``add character development'', etc. We used 34,000 preference data samples for fine-tuning the AD LLM to understand the downstream task of choosing the next story direction, and we used 25,000 samples to train the preference model using DPO. In the DPO dataset, we noticed an imbalance in the distribution of chosen actions by GPT-4. In \cref{fig:ogacfreq}, we can see the substantial difference in the number of stories for which ``add suspense'' was selected compared to other options. This observation implies that GPT-4 has an inherent bias while selecting actions to continue the story.

\begin{figure}[!ht]
    \centering
\includegraphics[trim={6cm 0 6cm 0}, clip, width=3.2in]{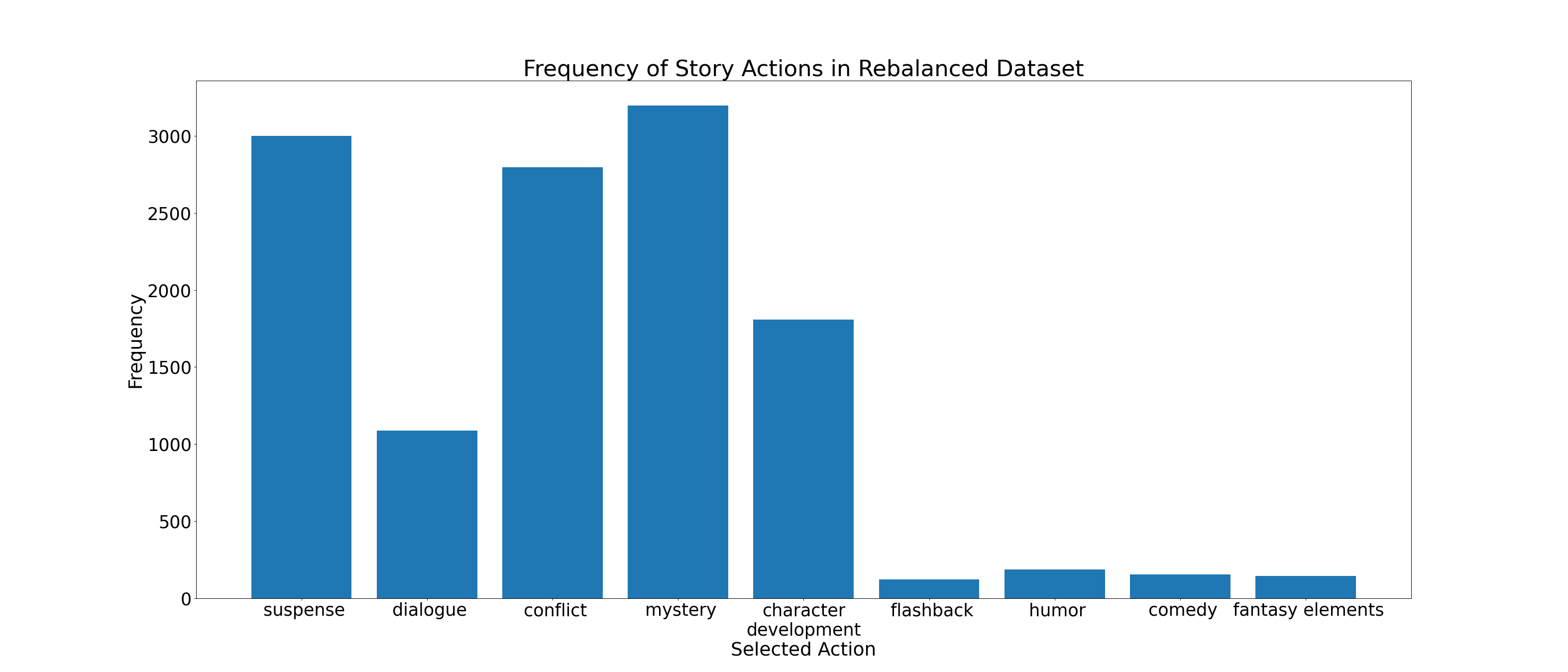}
    \caption{\textbf{Rebalanced  Distribution of Actions.} After our rebalancing procedure, we observe a more uniform distribution among the top 5 actions chosen. \textit{Note: actions chosen with frequency less than 100 not shown.}}
    \label{fig:newacfreq}
\end{figure}

In order to mitigate this effect, we generated more preference data from GPT-4, but this time, we removed the option to add suspense to the story. This would force GPT-4 to focus on other actions as well, resulting in a more spread out distribution of actions. After generating the new data, we took a random sample of 3,000 prompts from the original preference dataset with ``add suspense'' as the chosen action and merged it with our new dataset. In \cref{fig:newacfreq}, we can view the new distribution of story actions and notice that it is much more spread out, allowing for more variability in future story directions.

We collected three different datasets for SFT, DPO, and evaluation. Rebalancing was only done on the DPO dataset. Due to constraints with the GPT-4 API, we were unable to generate enough data for rebalancing the SFT dataset. However, it is worth noting that the SFT process allows our model to better understand the downstream task, but the DPO procedure is more critical for generating a preference model that produces useful results as shown in later experiments.

\subsection{Training}

For our AD LLM training, we first used a dataset of long stories to fine-tune our model to process long-context sequences. Then, we use a separate preference dataset collected for SFT to fine-tune our base AD LLM. We used approximately 34,000 ranking samples for SFT, and we trained the model to predict the next best action given the initial story state. We fine-tuned Llama-2-7B on this dataset for 5300 steps, with a mini-batch size of 1 and 64 gradient accumulation steps using 8 $\times$ A100 80GB GPUs (so one step processes 64 stories, and 530 steps is about one epoch). Completing the SFT process for each model required about 36 hours. We used the LongLoRA \cite{chen2023longlora} approach with Flash Attention 2.0 \cite{dao2023flashattention2} for SFT to enable fast fine-tuning on limited compute. We used the AdamW optimizer \cite{loshchilov2019decoupled} with $\beta_1 = 0.9, \beta_2 = 0.95$, a learning rate of 3e-5, and 30 warm-up steps with a constant learning rate scheduler.



We used DPO to train a preference model on two SFT model checkpoints, which were trained for 2650 and 5300 steps, respectively. The DPO training ran for 1000 steps for each model on approximately 25,000 samples of our preference dataset. We used a learning rate of 5e-4 with an AdamW optimizer and cosine annealing scheduler, both on default settings of $\beta_1=0.9, \beta_2=0.999$. We also used LoRA in our DPO training for both checkpoints, with $\alpha = 16$, $r=8$, and a dropout of 0.05. We conducted the DPO training using the Hugging Face Transformers Reinforcement Learning (TRL) \cite{vonwerra2022trl} library in a similar setting as SFT with 8 $\times$ A100 80GB GPUs but with a mini-batch size of 1 and 8 gradient accumulation steps. Each DPO training required approximately 12 hours with this setup on the rebalanced preference dataset, and we checkpointed our model at every 100 training steps. DPO for both checkpoints displayed convergence after approximately 800 steps of training.

\subsection{Inference}

\vspace{-0.2cm}
\begin{figure}[!ht]
    \centering
    \includegraphics[width=\columnwidth,trim={5cm 3cm 5cm 3cm},clip]{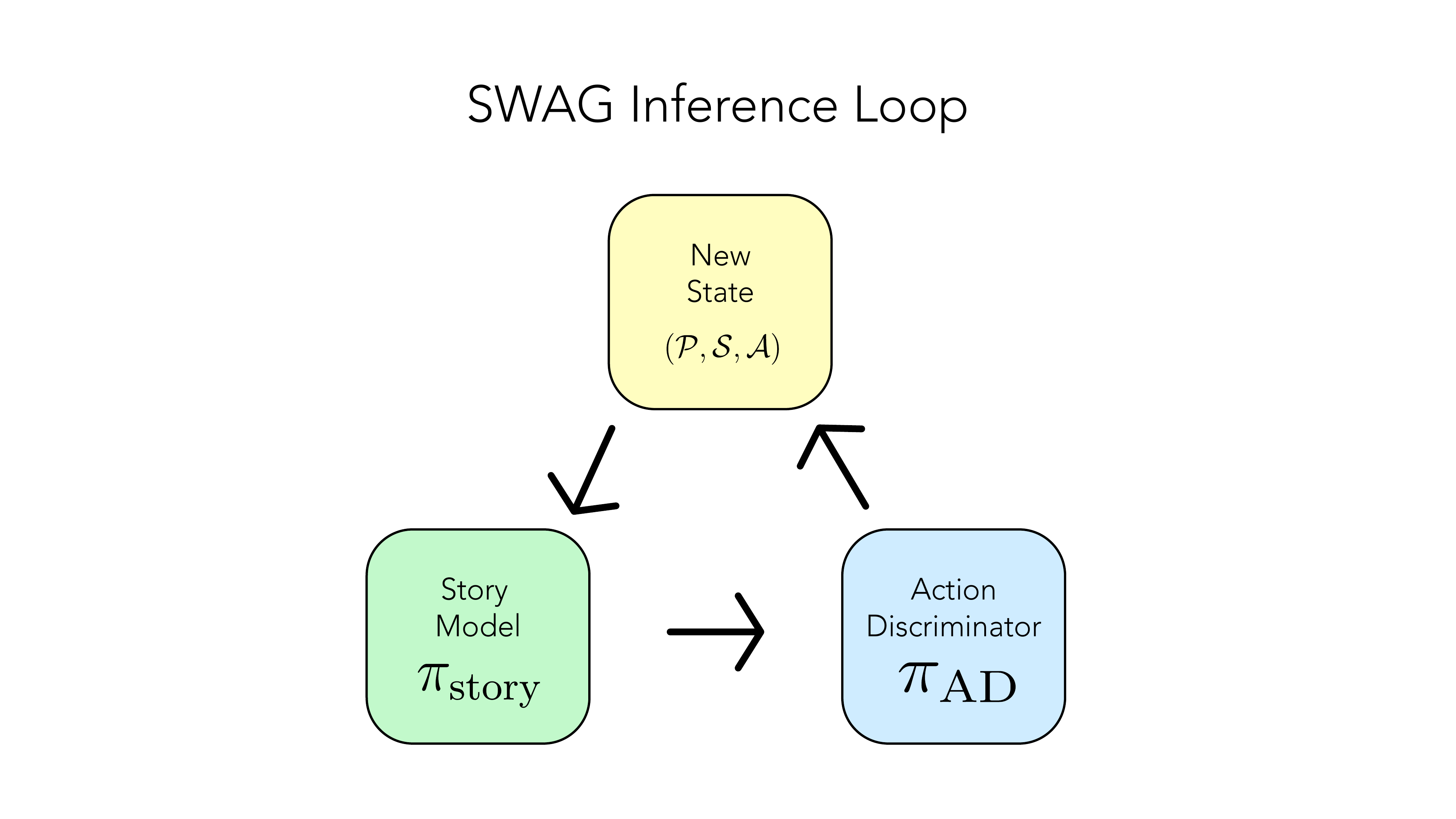}\caption{\textbf{\methodshort{} Inference Loop}. After sampling a story prompt and generating the initial paragraph, we pass the story state to our AD LLM to generate the next story action. The new state is passed back to the story model, and the process is repeated till a complete story is generated.}
    \label{fig:swaginference}
\end{figure}

Our inference pipeline requires two models: the action discriminator $\pi_{\text{AD}}$ and the story generation model $\pi_{\text{story}}$. We create a feedback loop between these two models to generate our story, depicted in Figure \ref{fig:swaginference}.

For our experiments, we evaluated the performance of different combinations of $\pi_{\text{AD}}$ and $\pi_{\text{story}}$ across a set of test story prompts. For each story prompt $\mathcal{P}$, we ask  $\pi_{\text{story}}$ to write the initial paragraph, and then, with this initial story state $(\mathcal{P}, \mathcal{S})$, we instruct $\pi_{\text{AD}}$ to select the optimal action for the subsequent paragraph.

In the action discriminator model $\pi_{\text{AD}}$ ablation, we used our own fine-tuned and aligned Llama-2-7B and Mistral-7B AD LLMs and GPT-4-Turbo. For the story generation model $\pi_{\text{story}}$ ablation, we used the base Llama-2-7B, Mistral-7B, GPT-3.5-Turbo, and GPT-4-Turbo models. For our open-source model generations, we also compare the performance when using a $\pi_{\text{AD}}$ that was tuned with a different base model as $\pi_{\text{story}}$.

To analyze the baseline performance for story generation, we generated stories with each $\pi_{\text{story}}$ by giving an initial story prompt and repeatedly prompting it to continue the story. The results of these end-to-end (E2E) generation ablations are shown in Table~\ref{table:turbo-evaluation}.

Finally, we analyze if our $\pi_{\text{AD}}$ models are better in choosing actions than a random selection. Our AD LLMs, trained using DPO, had a choice of only 30 actions during SFT and DPO. Using these 30 actions, we generated stories from the base Llama-2-7B and Mistral-7B models using our \methodshort{} pipeline. However, in this ablation, we replaced $\pi_{\text{AD}}$ and instead selected an action randomly from the list for each step of the loop.

\subsection{Summary of Ablations}
We perform several ablations on $\pi_{\text{story}}$ and $\pi_{\text{AD}}$ to test the performance of our algorithm. Specifically, we run pairwise comparisons between different combinations of $\pi_{\text{story}}$ and $\pi_{\text{AD}}$ models to gauge the quality of stories generated by \methodshort{}.

In the $\pi_{\text{story}}$ ablation study, we test different models to generate the story with a fixed $\pi_{\text{AD}}$. We run the \methodshort{} inference loop with several open-source and closed-source LLMs as $\pi_{\text{story}}$. This ablation provides insight into the level of improvement in story quality from different base models.

In the $\pi_{\text{AD}}$ ablation study, we test different models to generate the next story action with a fixed $\pi_{\text{story}}$. We trained two different AD LLMs for this ablation with the same SFT and DPO preference datasets.

To test \methodshort{} on closed-source LLMs, we also set up our inference pipeline with GPT-3.5-Turbo and GPT-4-Turbo. Here, we simply set GPT-3.5-Turbo and GPT-4-Turbo to be both $\pi_{\text{AD}}$ and $\pi_{\text{story}}$ in the \methodshort{} feedback loop. With these experiments, we aim to show the effectiveness of \methodshort{} even without fine-tuning an AD as a preference model. 

\subsection{Human Evaluation}
Our human evaluation setup is heavily inspired by \cite{zhu2023endtoend}. We run human evaluations comparing stories generated by various methods across three aspects: interesting-ness, surprise, and coherence. For each of 12 pairwise comparisons of two methods, we ask Surge AI workers to answer three preference questions about 50 pairs of stories generated by the methods we compare. We display the preference questions in Table \ref{human-eval-qs}, where each question corresponds to an aspect of story quality. We display our human annotation results in Figures 5-8.

\begin{table}[!h]
\begin{tabularx}{\columnwidth}{cX}
\toprule
\textbf{Q1} & Which story plot is more \textbf{interesting} to you overall? \\
\midrule
\textbf{Q2} & Which story created more \textbf{suspense} and \textbf{surprise}? \\
\midrule
\textbf{Q3} & Which story is more \textbf{coherent} and \textbf{consistent} in terms of plot structure?
\\
\bottomrule
\end{tabularx}
\caption{\label{human-eval-qs}Three questions asked to human annotators for 50 comparison story plot pairs.}
\end{table}

\begin{figure}[H]
    \label{fig:against-gpt-3.5-e2e}
    \vspace{-1em}
    \centering    
    \includegraphics[width=\linewidth]{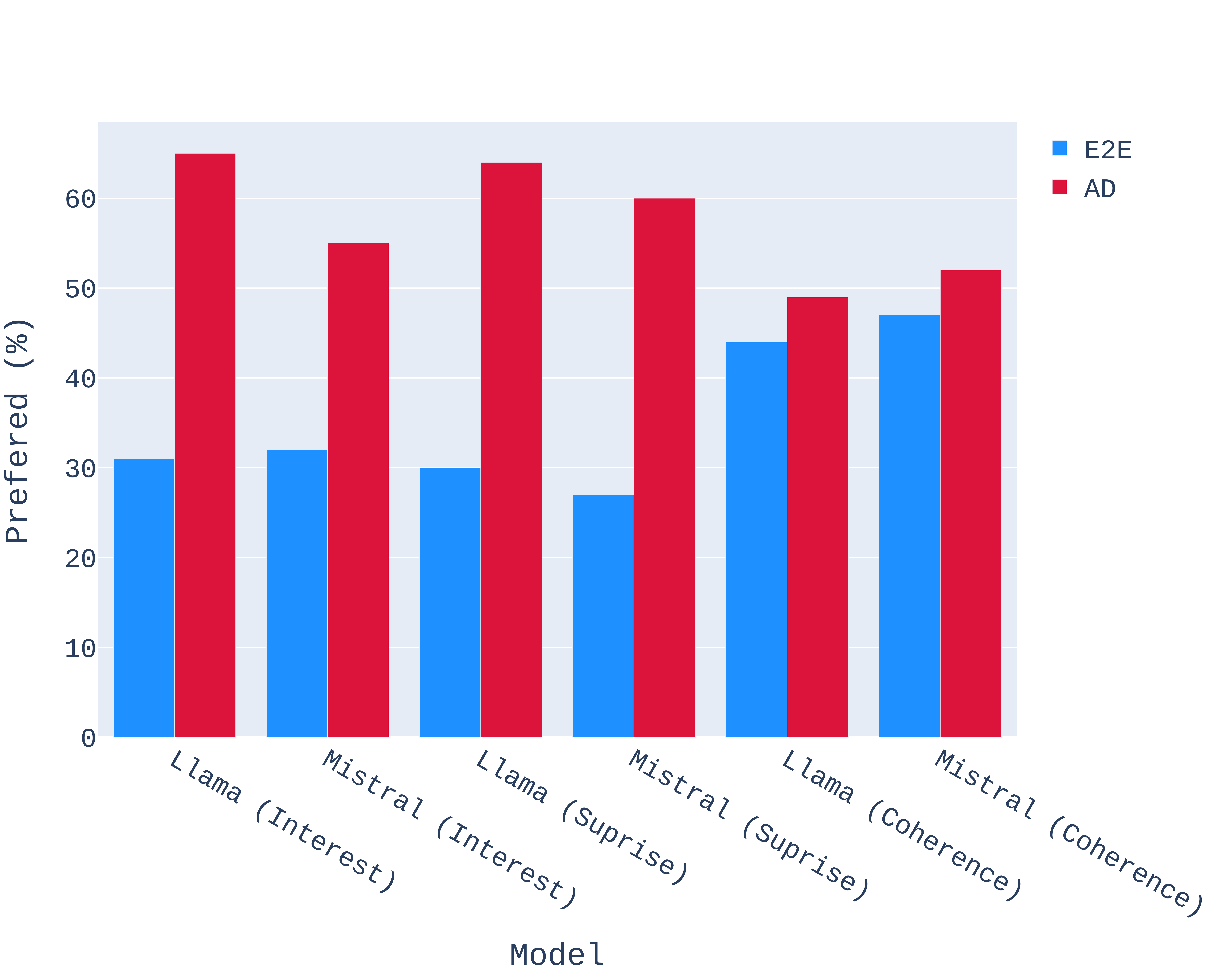}
    \vspace{-20px}
    \caption{Comparing SWAG with only Llama-2-7B or Mistral-7B (as both AD and story generator) against GPT-3.5-Turbo E2E on human evaluation data. The win-rate is calculated by averaging wins, losses, and ties. We count win as a score of 1, tie as a score of 0.5, and loss as a score of 0. Notably, we observe that using SWAG with smaller open-source models outperforms the larger GPT-3.5 model.}
\end{figure}

\begin{figure}[H]
    \label{fig:open-model-ad-improvement}
    \vspace{-1em}
    \centering
    \includegraphics[width=\linewidth]{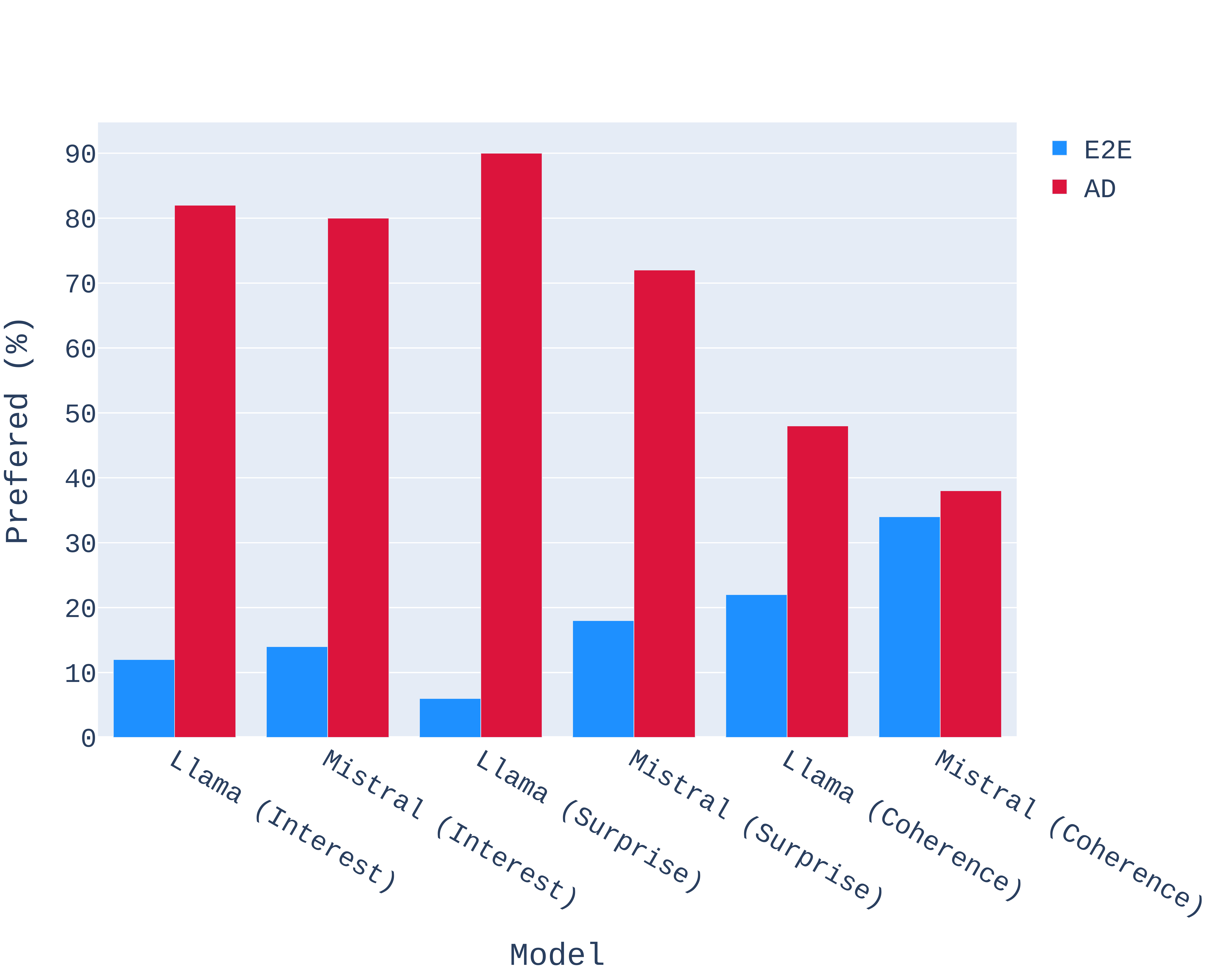}
    \vspace{-20px}
    \caption{Preferred rates between E2E and pure SWAG for Llama-2-7B and Mistral-7B on human evaluation data. We observe that for these open-source models, applying SWAG improves story generation outputs when compared to end-to-end generation.}
    \vspace{-10px}
\end{figure}

\begin{figure}[!h]
    \label{fig:closed-model-ad-improvement}
    \vspace{-1em}
    \centering
    \includegraphics[width=\linewidth]{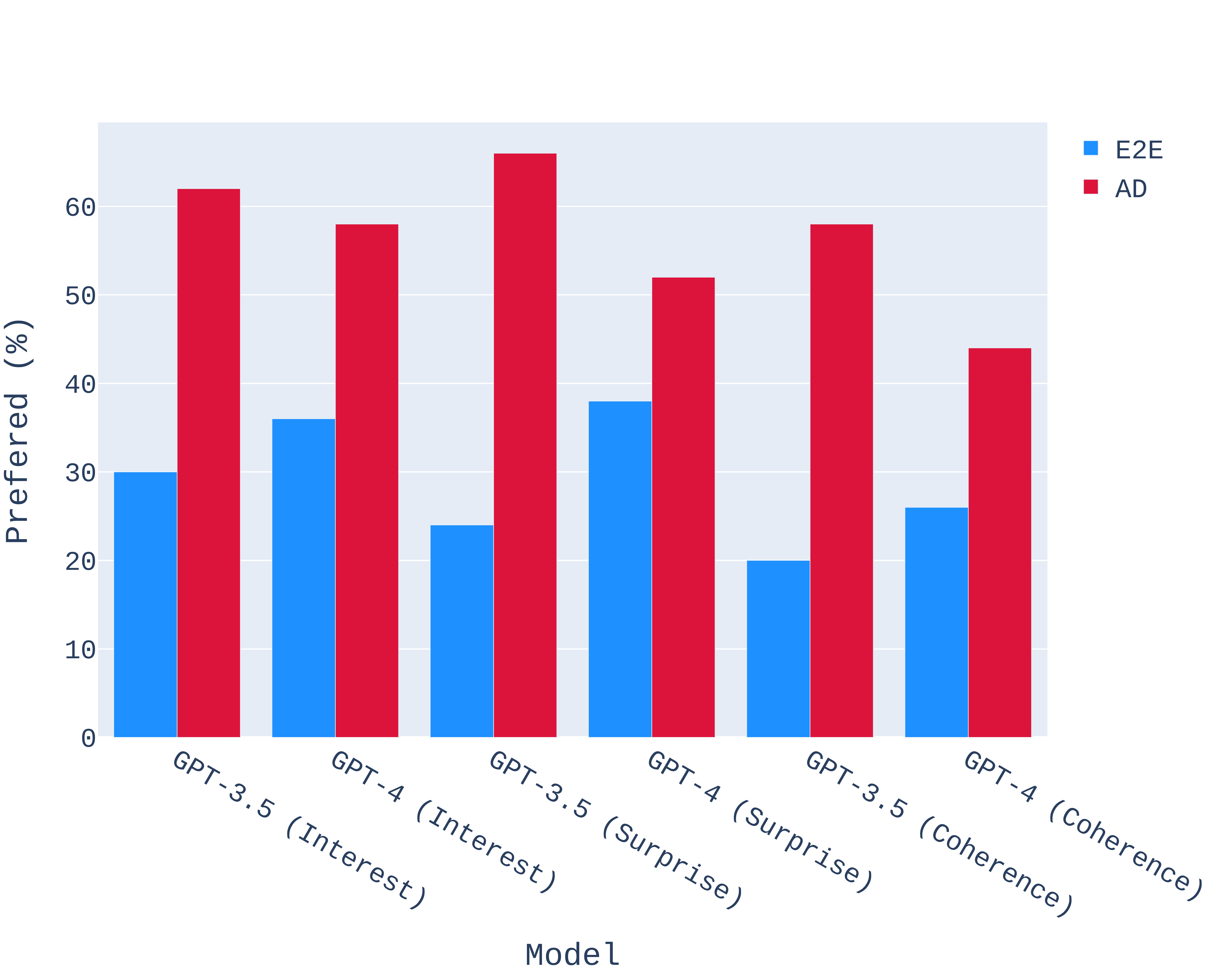}
    \vspace{-20px}
    \caption{Preferred rate between E2E and pure SWAG for GPT-3.5-Turbo and GPT-4-Turbo on human evaluation data. We observe that for these closed-source models, applying SWAG improves story generation outputs when compared to end-to-end generation.}
    \vspace{-10px}
\end{figure}

\begin{figure}[!h]
    \label{fig:random_vs_ad_chart}
    \vspace{-1em}
    \centering
    \includegraphics[width=\linewidth]{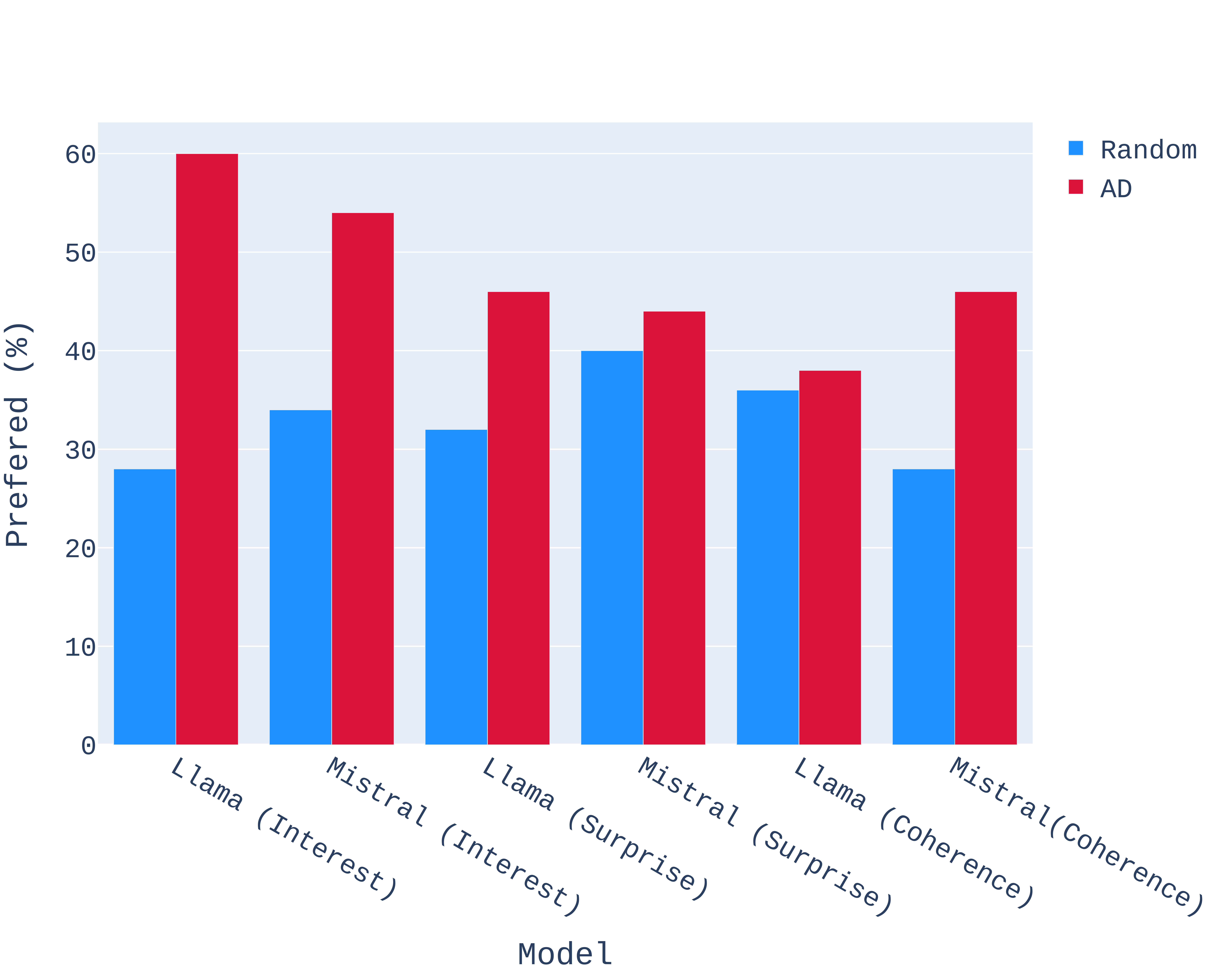}
    \vspace{-20px}
    \caption{Preferred rate between a random action AD and a model AD for Llama-2-7B and Mistral-7B on human evaluation data. We find that within SWAG, a LLM AD outperforms a random action AD.}
    \vspace{-10px}
\end{figure}

\newpage

\subsection{Machine (GPT-4-Turbo) Evaluation}
Recent developments in open-ended benchmarks shows promising results in evaluating LLM's response, with increasing utilization of GPT-4 in place of human judges, such as MT-Bench \cite{zheng2023judging} and AlpacaEval \cite{dubois2023alpacafarm}. Employing a similar strategy, we conduct evaluations with GPT-4-Turbo as a judge to pairwise compare two stories and pick the more interesting, engaging, and consistent story or a tie. The system prompt can be found in Appendix~\ref{app:system-prompt}. We evaluated several open and proprietary variants of SWAG against different baselines (using a random action AD, GPT-3.5-Turbo end-to-end generation, etc.), with results presented in Table~\ref{table:turbo-evaluation}.
\\
\begin{table}[h]
\setlength{\tabcolsep}{4pt}
\centering
\vspace{-1em}
\footnotesize
{\small
\begin{tabular}{lrrrr}
\toprule
SWAG vs E2E  & Win-Rate & SWAG  & E2E & Tie \\
\midrule
Mistral-7B      & \textbf{68.0\%} & 58 & 22 & 20 \\
Llama-2-7B      & \textbf{54.5\%} & 47 & 38 & 15 \\
GPT-3.5-Turbo   & \textbf{77.5\%} & 66 & 8  & 23 \\
GPT-4-Turbo     & \textbf{61.5\%} & 49 & 24 & 25 \\
\midrule
SWAG vs Rand AD & Win-Rate & SWAG & Rand AD & Tie \\
\midrule
Llama-2-7B   & \textbf{53.0\%} & 45 & 39 & 16 \\
Mistral-7B   & \textbf{67.5\%}   & 61 & 26 & 13 \\
\midrule
SWAG vs GPT-3.5       & Win-Rate & SWAG  & GPT-3.5 & Tie \\
\midrule
Mistral-7B      & 19.5\%      & 11 & 72 & 17 \\
Llama-2-7B      & 31.0\%    & 19 & 57 & 24 \\
\midrule
E2E vs GPT-3.5       & Win-Rate & E2E  & GPT-3.5 & Tie \\
\midrule
Mistral-7B      & 9.5\%      & 3 & 84 & 13 \\
Llama-2-7B      & 23.5\%    & 14 & 67 & 19 \\
\bottomrule
\end{tabular}
}
\caption{\label{table:turbo-evaluation}Evaluation results of pairwise comparisons between \methodshort{} (with LLM AD) vs. baselines, with GPT-4-Turbo as the judge. The win-rate is calculated by averaging wins, losses, and ties. We count win as a score of 1, tie as a score of 0.5, and loss as a score of 0.}
\vspace{-0.5em}
\end{table}



\section{Discussion}

\subsection{Machine Evaluation Results}
 Table~\ref{table:turbo-evaluation} displays the pairwise evaluation results using GPT-4-Turbo as a judge. The win-rate column specifies the percentage of stories generated by \methodshort{} that were preferred by the LM judge in the comparison. For the AD vs. Random comparisons, GPT-4 preferred Llama-2-7B and Mistral-7B with \methodshort{} over using randomly selected actions. This shows that the AD LLM in \methodshort{} provides useful signals to the story generation LLM for guiding the story direction. 

In the AD vs. E2E comparisons, \methodshort{} outperforms the E2E approach across all models. We note a significantly large win-rate in \methodshort{} results for Mistral-7B, GPT-3.5-Turbo, and GPT-4-Turbo and a slightly higher win-rate than E2E with Llama-2-7B. This indicates that \methodshort{} is greatly improves story engagement compared to generating long-form stories with no guidance.

The results across the ablations exhibit the effectiveness of \methodshort{} and how a simple feedback loop improves content quality in stories. In each evaluation, GPT-4-Turbo provides reasoning for its story preference ranking. The stories generated with \methodshort{} are consistently rated to have better suspense, surprise, and engagement. Examples of GPT-4-Turbo's reasoning can be seen in Appendix~\ref{app:turbo-judgment-examples}.



\subsection{Human Evaluation Results}

We then evaluate these stories once again in terms of interesting-ness, surprise, and coherence with humans as the judge. The human evaluators were specifically asked to rate each aspect separately by answering the questions in Table~\ref{human-eval-qs}. We provide the full results in Appendix \ref{full_human_eval}. For both open-source and closed-source models, \methodshort{} produces stories that overwhelmingly beat their E2E counterparts. In particular, we highlight that both \methodshort{} Llama-2-7B's stories and \methodshort{} Mistral-7B's stories were significantly more preferred over GPT-3.5-Turbo's stories along interest and surprise while being equivalent in coherence; see Table \ref{human-eval-2} in Appendix \ref{full_human_eval} for more details.

Comparing GPT-4-Turbo and human evaluation, AD consistently outperform its baseline regardless of judges, demonstrating SWAG's effectiveness. However, the gap in preferences is greater in human evaluation in comparison to GPT-4-Turbo as judge. As shown in Table~\ref{table:turbo-evaluation} and Table~\ref{human-eval-2}, there is a significant difference in preferences on pairwise comparisons between open source AD LLMs and GPT-3.5-Turbo, with only 14\% of Llama-2-7B AD being preferred over GPT-3.5-Turbo when GPT-4-Turbo is judge, while over 50\% of Llama-2-7B AD being preferred across the 3 aspects. This is most likely due to GPT-4-Turbo inherent bias towards GPT-3.5-Turbo while human evaluators does not have a bias towards any particular LLM. These inconsistencies between GPT-4-Turbo and human judges reveal that even the strongest propriety models continue to lag behind human evaluators in terms of quality and trustworthiness.  

\subsection{Extensions}

Beyond generating the story automatically using \methodshort{}, users can also intervene in the story generation process. Our method can be ``paused'' at any time, after which a human can continue writing the story or even collaborate back-and-forth with the story model via \methodshort{}. We are excited to explore new forms of human-LLM interaction as automated generation capabilities progress.

To further customize the \methodshort{} inference loop, the user can also tailor the list of actions for the AD LLM to their own needs. For example, if a user would like their AD LLM to specialize in directing stories that focus on a specific genre like horror, they can add actions that better fit this theme. The flexibility to choose actions allows \methodshort{} to be a versatile system for a wide variety of content generation tasks across various genres. 

Based on our experiments and evaluations, we believe that our results could be further improved given more fine-grained actions during SFT and DPO training and inference time. Fine-grained actions would enable consistent control and can add depth and complexity to stories to increase engagement with the reader. Using more detailed actions can lead to richer narratives by allowing for more nuanced character development, plot twists, and detailed settings. 

Another avenue for improving results could be done by generating actions at "test-time", i.e. by prompting the story LLM to generate a set of actions given the current state of the story. The AD LLM would then select an action from this generated set. In our preliminary experiments, we found this to be compelling approach since it allows for far more fine-grain action generation which can directly tie in story elements that otherwise couldn't be accounted for when generating a fixed action bank.

\section{Conclusion}

This paper proposes \methodshort{}, a simple feedback-based framework for creative story generation. The fine-tuned action discriminator LLM enables more interesting and exciting plot development with little to no sacrifice in coherence or consistency. Both machine and human evaluation exemplify our method's effectiveness compared to SoTA end-to-end generation methods, even with the strongest closed-source models. We anticipate that our contribution will further advancements in content generation, particularly through the lens of iterative feedback mechanisms.


\section{Limitations and Ethical Concerns}


\subsection{Limitations}
Due to compute restraints, we were only able to use DPO for AD LLM alignment. DPO is much more lightweight than PPO as it is an offline RL algorithm. However, it is possible that with the online sampling process of PPO and with a strong reward model, we would be able to achieve better results. We also would have preferred to increase the scope of our ablations, potentially experimenting with a greater variety open-source and closed-source models and sampling from a larger set of diverse and fine-grained actions at test-time. 


For our evaluations, we were only able to generate machine evaluations on 100 test story prompts and human evaluations on 50 test story prompts due to resource constraints. Evaluating on a larger set of stories, especially for machine evaluation, would give us better insight into the quality of the stories generated by \methodshort{}. We also conducted the evaluations before the release of stronger models such as Llama-3 and GPT-4o, which could have improved our results. However, due to budget constraints, we were unable to run another set of evaluations for new models. 

Although the scope of our ablations does not fully encapsulate all the possibilities of \methodshort{}, our work displays the effectiveness of generation with iterative feedback and provides an initial perspective on automating controlled content generation from LLMs. 

\subsection{Ethical Concerns}
We do not foresee any major immediate ethical or societal impacts resulting from our method. Our method focuses on providing additional control over content generation from LLMs. We do acknowledge it is possible to use our method to tune a model to copy the writing style of another author. However, this would require a large amount of labeled data to steer the LLM in this direction. Finally, we do see potential in using our method in processes that require iterative feedback such as planning trajectories for robotics. These applications would need to ensure that the actions dataset used to apply our method meets specific ethical criteria.

\bibliography{custom}

\begin{thebibliography}{37}
\providecommand{\natexlab}[1]{#1}

\bibitem[{Alabdulkarim et~al.(2021{\natexlab{a}})Alabdulkarim, Li, and Peng}]{alabdulkarim2021automatic}
Amal Alabdulkarim, Siyan Li, and Xiangyu Peng. 2021{\natexlab{a}}.
\newblock \href {https://arxiv.org/abs/2102.12634} {Automatic story generation: Challenges and attempts}.
\newblock \emph{Preprint}, arXiv:2102.12634.

\bibitem[{Alabdulkarim et~al.(2021{\natexlab{b}})Alabdulkarim, Li, Martin, and Riedl}]{alabdulkarim2021goaldirected}
Amal Alabdulkarim, Winston Li, Lara~J. Martin, and Mark~O. Riedl. 2021{\natexlab{b}}.
\newblock \href {https://arxiv.org/abs/2112.08593} {Goal-directed story generation: Augmenting generative language models with reinforcement learning}.
\newblock \emph{Preprint}, arXiv:2112.08593.

\bibitem[{Bai et~al.(2022)Bai, Jones, Ndousse, Askell, Chen, DasSarma, Drain, Fort, Ganguli, Henighan, Joseph, Kadavath, Kernion, Conerly, El-Showk, Elhage, Hatfield-Dodds, Hernandez, Hume, Johnston, Kravec, Lovitt, Nanda, Olsson, Amodei, Brown, Clark, McCandlish, Olah, Mann, and Kaplan}]{bai2022training}
Yuntao Bai, Andy Jones, Kamal Ndousse, Amanda Askell, Anna Chen, Nova DasSarma, Dawn Drain, Stanislav Fort, Deep Ganguli, Tom Henighan, Nicholas Joseph, Saurav Kadavath, Jackson Kernion, Tom Conerly, Sheer El-Showk, Nelson Elhage, Zac Hatfield-Dodds, Danny Hernandez, Tristan Hume, Scott Johnston, Shauna Kravec, Liane Lovitt, Neel Nanda, Catherine Olsson, Dario Amodei, Tom Brown, Jack Clark, Sam McCandlish, Chris Olah, Ben Mann, and Jared Kaplan. 2022.
\newblock \href {https://arxiv.org/abs/2204.05862} {Training a helpful and harmless assistant with reinforcement learning from human feedback}.
\newblock \emph{Preprint}, arXiv:2204.05862.

\bibitem[{Brown et~al.(2020)Brown, Mann, Ryder, Subbiah, Kaplan, Dhariwal, Neelakantan, Shyam, Sastry, Askell, Agarwal, Herbert{-}Voss, Krueger, Henighan, Child, Ramesh, Ziegler, Wu, Winter, Hesse, Chen, Sigler, Litwin, Gray, Chess, Clark, Berner, McCandlish, Radford, Sutskever, and Amodei}]{gpt3}
Tom~B. Brown, Benjamin Mann, Nick Ryder, Melanie Subbiah, Jared Kaplan, Prafulla Dhariwal, Arvind Neelakantan, Pranav Shyam, Girish Sastry, Amanda Askell, Sandhini Agarwal, Ariel Herbert{-}Voss, Gretchen Krueger, Tom Henighan, Rewon Child, Aditya Ramesh, Daniel~M. Ziegler, Jeffrey Wu, Clemens Winter, Christopher Hesse, Mark Chen, Eric Sigler, Mateusz Litwin, Scott Gray, Benjamin Chess, Jack Clark, Christopher Berner, Sam McCandlish, Alec Radford, Ilya Sutskever, and Dario Amodei. 2020.
\newblock \href {https://arxiv.org/abs/2005.14165} {Language models are few-shot learners}.
\newblock \emph{CoRR}, abs/2005.14165.

\bibitem[{Castricato et~al.(2022)Castricato, Havrilla, Matiana, Pieler, Ye, Yang, Frazier, and Riedl}]{castricato2022robust}
Louis Castricato, Alexander Havrilla, Shahbuland Matiana, Michael Pieler, Anbang Ye, Ian Yang, Spencer Frazier, and Mark Riedl. 2022.
\newblock \href {https://arxiv.org/abs/2210.07792} {Robust preference learning for storytelling via contrastive reinforcement learning}.
\newblock \emph{Preprint}, arXiv:2210.07792.

\bibitem[{Chang et~al.(2023)Chang, Brantley, Ramamurthy, Misra, and Sun}]{chang2023learning}
Jonathan~D. Chang, Kiante Brantley, Rajkumar Ramamurthy, Dipendra Misra, and Wen Sun. 2023.
\newblock \href {https://arxiv.org/abs/2306.11816} {Learning to generate better than your llm}.
\newblock \emph{Preprint}, arXiv:2306.11816.

\bibitem[{Charniak(2004)}]{charniak2004storycomp}
Eugene Charniak. 2004.
\newblock Toward a model of children's story comprehension.

\bibitem[{Chen et~al.(2023)Chen, Qian, Tang, Lai, Liu, Han, and Jia}]{chen2023longlora}
Yukang Chen, Shengju Qian, Haotian Tang, Xin Lai, Zhijian Liu, Song Han, and Jiaya Jia. 2023.
\newblock \href {https://arxiv.org/abs/2309.12307} {Longlora: Efficient fine-tuning of long-context large language models}.
\newblock \emph{Preprint}, arXiv:2309.12307.

\bibitem[{Chung et~al.(2022)Chung, Kim, Yoo, Lee, Adar, and Chang}]{chung2022talebrush}
John Joon~Young Chung, Wooseok Kim, Kang~Min Yoo, Hwaran Lee, Eytan Adar, and Minsuk Chang. 2022.
\newblock Talebrush: Sketching stories with generative pretrained language models.
\newblock In \emph{Proceedings of the 2022 CHI Conference on Human Factors in Computing Systems}, pages 1--19.

\bibitem[{Coenen et~al.(2021)Coenen, Davis, Ippolito, Reif, and Yuan}]{coenen2021wordcraft}
Andy Coenen, Luke Davis, Daphne Ippolito, Emily Reif, and Ann Yuan. 2021.
\newblock \href {https://arxiv.org/abs/2107.07430} {Wordcraft: a human-ai collaborative editor for story writing}.
\newblock \emph{Preprint}, arXiv:2107.07430.

\bibitem[{Dao(2023)}]{dao2023flashattention2}
Tri Dao. 2023.
\newblock \href {https://arxiv.org/abs/2307.08691} {Flashattention-2: Faster attention with better parallelism and work partitioning}.
\newblock \emph{Preprint}, arXiv:2307.08691.

\bibitem[{Dubois et~al.(2023)Dubois, Li, Taori, Zhang, Gulrajani, Ba, Guestrin, Liang, and Hashimoto}]{dubois2023alpacafarm}
Yann Dubois, Xuechen Li, Rohan Taori, Tianyi Zhang, Ishaan Gulrajani, Jimmy Ba, Carlos Guestrin, Percy Liang, and Tatsunori~B Hashimoto. 2023.
\newblock Alpacafarm: A simulation framework for methods that learn from human feedback.
\newblock \emph{arXiv preprint arXiv:2305.14387}.

\bibitem[{Fan et~al.(2018)Fan, Lewis, and Dauphin}]{fan2018hierarchical}
Angela Fan, Mike Lewis, and Yann Dauphin. 2018.
\newblock \href {https://arxiv.org/abs/1805.04833} {Hierarchical neural story generation}.
\newblock \emph{Preprint}, arXiv:1805.04833.

\bibitem[{Goldfarb-Tarrant et~al.(2019)Goldfarb-Tarrant, Feng, and Peng}]{goldfarb-tarrant-etal-2019-plan}
Seraphina Goldfarb-Tarrant, Haining Feng, and Nanyun Peng. 2019.
\newblock \href {https://doi.org/10.18653/v1/N19-4016} {Plan, write, and revise: an interactive system for open-domain story generation}.
\newblock In \emph{Proceedings of the 2019 Conference of the North {A}merican Chapter of the Association for Computational Linguistics (Demonstrations)}, pages 89--97, Minneapolis, Minnesota. Association for Computational Linguistics.

\bibitem[{Jiang et~al.(2023)Jiang, Sablayrolles, Mensch, Bamford, Chaplot, de~las Casas, Bressand, Lengyel, Lample, Saulnier, Lavaud, Lachaux, Stock, Scao, Lavril, Wang, Lacroix, and Sayed}]{jiang2023mistral}
Albert~Q. Jiang, Alexandre Sablayrolles, Arthur Mensch, Chris Bamford, Devendra~Singh Chaplot, Diego de~las Casas, Florian Bressand, Gianna Lengyel, Guillaume Lample, Lucile Saulnier, Lélio~Renard Lavaud, Marie-Anne Lachaux, Pierre Stock, Teven~Le Scao, Thibaut Lavril, Thomas Wang, Timothée Lacroix, and William~El Sayed. 2023.
\newblock \href {https://arxiv.org/abs/2310.06825} {Mistral 7b}.
\newblock \emph{Preprint}, arXiv:2310.06825.

\bibitem[{Lester et~al.(2021)Lester, Al-Rfou, and Constant}]{lester2021power}
Brian Lester, Rami Al-Rfou, and Noah Constant. 2021.
\newblock The power of scale for parameter-efficient prompt tuning.
\newblock \emph{arXiv preprint arXiv:2104.08691}.

\bibitem[{Lin and Riedl(2021)}]{Lin2021PlugandBlendAF}
Zhiyu Lin and Mark~O. Riedl. 2021.
\newblock \href {https://api.semanticscholar.org/CorpusID:236470168} {Plug-and-blend: A framework for plug-and-play controllable story generation with sketches}.
\newblock In \emph{Artificial Intelligence and Interactive Digital Entertainment Conference}.

\bibitem[{Loshchilov and Hutter(2019)}]{loshchilov2019decoupled}
Ilya Loshchilov and Frank Hutter. 2019.
\newblock \href {https://arxiv.org/abs/1711.05101} {Decoupled weight decay regularization}.
\newblock \emph{Preprint}, arXiv:1711.05101.

\bibitem[{Martin et~al.(2017)Martin, Ammanabrolu, Wang, Singh, Harrison, Dhuliawala, Tambwekar, Mehta, Arora, Dass et~al.}]{martin2017improvisational}
Lara~J Martin, Prithviraj Ammanabrolu, Xinyu Wang, Shruti Singh, Brent Harrison, Murtaza Dhuliawala, Pradyumna Tambwekar, Animesh Mehta, Richa Arora, Nathan Dass, et~al. 2017.
\newblock Improvisational storytelling agents.
\newblock In \emph{Workshop on Machine Learning for Creativity and Design (NeurIPS 2017)}, volume~8.

\bibitem[{Mirowski et~al.(2022)Mirowski, Mathewson, Pittman, and Evans}]{mirowski2022cowriting}
Piotr Mirowski, Kory~W. Mathewson, Jaylen Pittman, and Richard Evans. 2022.
\newblock \href {https://arxiv.org/abs/2209.14958} {Co-writing screenplays and theatre scripts with language models: An evaluation by industry professionals}.
\newblock \emph{Preprint}, arXiv:2209.14958.

\bibitem[{Oatley(1995)}]{oatley-1995-book}
Keith Oatley. 1995.
\newblock \href {https://aclanthology.org/J95-4007} {Book reviews: The creative process: A computer model of storytelling and creativity}.
\newblock \emph{Computational Linguistics}, 21(4).

\bibitem[{OpenAI(2023)}]{OpenAI_GPT4_2023}
OpenAI. 2023.
\newblock \href {https://arxiv.org/abs/2303.08774} {Gpt-4 technical report}.
\newblock \emph{ArXiv}, abs/2303.08774.

\bibitem[{Peng et~al.(2022)Peng, Xie, Alabdulkarim, Kayam, Dani, and Riedl}]{peng2022guiding}
Xiangyu Peng, Kaige Xie, Amal Alabdulkarim, Harshith Kayam, Samihan Dani, and Mark~O. Riedl. 2022.
\newblock \href {https://arxiv.org/abs/2112.08596} {Guiding neural story generation with reader models}.
\newblock \emph{Preprint}, arXiv:2112.08596.

\bibitem[{Qin and Eisner(2021)}]{qin2021learning}
Guanghui Qin and Jason Eisner. 2021.
\newblock Learning how to ask: Querying lms with mixtures of soft prompts.
\newblock \emph{arXiv preprint arXiv:2104.06599}.

\bibitem[{Rafailov et~al.(2023)Rafailov, Sharma, Mitchell, Ermon, Manning, and Finn}]{rafailov2023direct}
Rafael Rafailov, Archit Sharma, Eric Mitchell, Stefano Ermon, Christopher~D. Manning, and Chelsea Finn. 2023.
\newblock \href {https://arxiv.org/abs/2305.18290} {Direct preference optimization: Your language model is secretly a reward model}.
\newblock \emph{Preprint}, arXiv:2305.18290.

\bibitem[{Rashkin et~al.(2020)Rashkin, Celikyilmaz, Choi, and Gao}]{rashkin2020plotmachines}
Hannah Rashkin, Asli Celikyilmaz, Yejin Choi, and Jianfeng Gao. 2020.
\newblock \href {https://arxiv.org/abs/2004.14967} {Plotmachines: Outline-conditioned generation with dynamic plot state tracking}.
\newblock \emph{Preprint}, arXiv:2004.14967.

\bibitem[{Shin et~al.(2020)Shin, Razeghi, au2, Wallace, and Singh}]{shin2020autoprompt}
Taylor Shin, Yasaman Razeghi, Robert L. Logan~IV au2, Eric Wallace, and Sameer Singh. 2020.
\newblock \href {https://arxiv.org/abs/2010.15980} {Autoprompt: Eliciting knowledge from language models with automatically generated prompts}.
\newblock \emph{Preprint}, arXiv:2010.15980.

\bibitem[{Tambwekar et~al.(2019)Tambwekar, Dhuliawala, Martin, Mehta, Harrison, and Riedl}]{Tambwekar_2019}
Pradyumna Tambwekar, Murtaza Dhuliawala, Lara~J. Martin, Animesh Mehta, Brent Harrison, and Mark~O. Riedl. 2019.
\newblock \href {https://doi.org/10.24963/ijcai.2019/829} {Controllable neural story plot generation via reward shaping}.
\newblock In \emph{Proceedings of the Twenty-Eighth International Joint Conference on Artificial Intelligence}, IJCAI-2019. International Joint Conferences on Artificial Intelligence Organization.

\bibitem[{Touvron et~al.(2023)Touvron, Martin, Stone, Albert, Almahairi, Babaei, Bashlykov, Batra, Bhargava, Bhosale, Bikel, Blecher, Ferrer, Chen, Cucurull, Esiobu, Fernandes, Fu, Fu, Fuller, Gao, Goswami, Goyal, Hartshorn, Hosseini, Hou, Inan, Kardas, Kerkez, Khabsa, Kloumann, Korenev, Koura, Lachaux, Lavril, Lee, Liskovich, Lu, Mao, Martinet, Mihaylov, Mishra, Molybog, Nie, Poulton, Reizenstein, Rungta, Saladi, Schelten, Silva, Smith, Subramanian, Tan, Tang, Taylor, Williams, Kuan, Xu, Yan, Zarov, Zhang, Fan, Kambadur, Narang, Rodriguez, Stojnic, Edunov, and Scialom}]{touvron2023llama}
Hugo Touvron, Louis Martin, Kevin Stone, Peter Albert, Amjad Almahairi, Yasmine Babaei, Nikolay Bashlykov, Soumya Batra, Prajjwal Bhargava, Shruti Bhosale, Dan Bikel, Lukas Blecher, Cristian~Canton Ferrer, Moya Chen, Guillem Cucurull, David Esiobu, Jude Fernandes, Jeremy Fu, Wenyin Fu, Brian Fuller, Cynthia Gao, Vedanuj Goswami, Naman Goyal, Anthony Hartshorn, Saghar Hosseini, Rui Hou, Hakan Inan, Marcin Kardas, Viktor Kerkez, Madian Khabsa, Isabel Kloumann, Artem Korenev, Punit~Singh Koura, Marie-Anne Lachaux, Thibaut Lavril, Jenya Lee, Diana Liskovich, Yinghai Lu, Yuning Mao, Xavier Martinet, Todor Mihaylov, Pushkar Mishra, Igor Molybog, Yixin Nie, Andrew Poulton, Jeremy Reizenstein, Rashi Rungta, Kalyan Saladi, Alan Schelten, Ruan Silva, Eric~Michael Smith, Ranjan Subramanian, Xiaoqing~Ellen Tan, Binh Tang, Ross Taylor, Adina Williams, Jian~Xiang Kuan, Puxin Xu, Zheng Yan, Iliyan Zarov, Yuchen Zhang, Angela Fan, Melanie Kambadur, Sharan Narang, Aurelien Rodriguez, Robert Stojnic, Sergey Edunov, and Thomas
  Scialom. 2023.
\newblock \href {https://arxiv.org/abs/2307.09288} {Llama 2: Open foundation and fine-tuned chat models}.
\newblock \emph{Preprint}, arXiv:2307.09288.

\bibitem[{von Werra et~al.(2020)von Werra, Belkada, Tunstall, Beeching, Thrush, Lambert, and Huang}]{vonwerra2022trl}
Leandro von Werra, Younes Belkada, Lewis Tunstall, Edward Beeching, Tristan Thrush, Nathan Lambert, and Shengyi Huang. 2020.
\newblock Trl: Transformer reinforcement learning.
\newblock \url{https://github.com/huggingface/trl}.

\bibitem[{Wang and Gordon(2023)}]{wang2023playing}
Timothy~S Wang and Andrew~S Gordon. 2023.
\newblock Playing story creation games with large language models: Experiments with gpt-3.5.
\newblock In \emph{International Conference on Interactive Digital Storytelling}, pages 297--305. Springer.

\bibitem[{Wei et~al.(2022)Wei, Wang, Schuurmans, Bosma, Ichter, Xia, Chi, Le, and Zhou}]{wei-chainofthought}
Jason Wei, Xuezhi Wang, Dale Schuurmans, Maarten Bosma, Brian Ichter, Fei Xia, Ed~Chi, Quoc Le, and Denny Zhou. 2022.
\newblock \href {https://doi.org/10.48550/ARXIV.2201.11903} {Chain-of-thought prompting elicits reasoning in large language models}.

\bibitem[{Wilmot and Keller(2021)}]{wilmot2021temporal}
David Wilmot and Frank Keller. 2021.
\newblock \href {https://arxiv.org/abs/2109.06807} {A temporal variational model for story generation}.
\newblock \emph{Preprint}, arXiv:2109.06807.

\bibitem[{Xu et~al.(2018)Xu, Ren, Zhang, Zeng, Cai, and Sun}]{xu2018skeletonbased}
Jingjing Xu, Xuancheng Ren, Yi~Zhang, Qi~Zeng, Xiaoyan Cai, and Xu~Sun. 2018.
\newblock \href {https://arxiv.org/abs/1808.06945} {A skeleton-based model for promoting coherence among sentences in narrative story generation}.
\newblock \emph{Preprint}, arXiv:1808.06945.

\bibitem[{Zheng et~al.(2023)Zheng, Chiang, Sheng, Zhuang, Wu, Zhuang, Lin, Li, Li, Xing, Zhang, Gonzalez, and Stoica}]{zheng2023judging}
Lianmin Zheng, Wei-Lin Chiang, Ying Sheng, Siyuan Zhuang, Zhanghao Wu, Yonghao Zhuang, Zi~Lin, Zhuohan Li, Dacheng Li, Eric~P. Xing, Hao Zhang, Joseph~E. Gonzalez, and Ion Stoica. 2023.
\newblock \href {https://arxiv.org/abs/2306.05685} {Judging llm-as-a-judge with mt-bench and chatbot arena}.
\newblock \emph{Preprint}, arXiv:2306.05685.

\bibitem[{Zhu et~al.(2023)Zhu, Cohen, Wang, Yang, Yang, Jiao, and Tian}]{zhu2023endtoend}
Hanlin Zhu, Andrew Cohen, Danqing Wang, Kevin Yang, Xiaomeng Yang, Jiantao Jiao, and Yuandong Tian. 2023.
\newblock \href {https://arxiv.org/abs/2310.08796} {End-to-end story plot generator}.
\newblock \emph{Preprint}, arXiv:2310.08796.

\bibitem[{Zou et~al.(2021)Zou, Yin, Zhong, Yang, Yang, and Tang}]{zou-2021-inverse}
Xu~Zou, Da~Yin, Qingyang Zhong, Hongxia Yang, Zhilin Yang, and Jie Tang. 2021.
\newblock \href {https://doi.org/10.1145/3447548.3467418} {Controllable generation from pre-trained language models via inverse prompting}.
\newblock In \emph{Proceedings of the 27th ACM SIGKDD Conference on Knowledge Discovery \& Data Mining}, KDD '21, page 2450–2460, New York, NY, USA. Association for Computing Machinery.

\end{thebibliography}

\newpage
\appendix
\onecolumn

\lstset{
    basicstyle=\ttfamily,
    breaklines=true,
}

\section{Prompts}
\subsection{AD LLM Prompt}
\label{app:AD-prompt}
\begin{tcolorbox}
\vspace{-1em}
\begin{lstlisting}
Here is a story prompt: {story_prompt}

Here is the story so far: {story}

Here is a set of actions: {actions}.

Based on the current story, choose the best action for the next paragraph.
Only output the action you chose without any quotation marks.
\end{lstlisting}
\vspace{-1em}
\end{tcolorbox}

\subsection{Story Model Prompt}
\label{app:story-prompt}
\begin{tcolorbox}
\vspace{-1em}
\begin{lstlisting}
Here is a story prompt: {story_prompt}

Here is the story so far: {story}

Here is an action for the next paragraph of the story: {action}. 

Write the next paragraph of the story such that it uses the given action.
New paragraph:
\end{lstlisting}
\vspace{-1em}
\end{tcolorbox}

\subsection{System Prompt for Evaluation}
\label{app:system-prompt}
\begin{tcolorbox}
\texttt{Please act as an impartial judge and evaluate the quality of the stories generated by two AI models. The two stories have the same premise. You should choose the stories that are more engaging and interesting, have better suspense and surprise, and are consistent and straightforward. Your evaluation should focus on which story is more interesting and engaging overall and which story created more suspense or surprise while remaining consistent with the initial story prompt. Do not evaluate the stories based on whether or not they are complete, have a clear resolution, have a larger scope, have more variety, or are more unpredictable. Only evaluate them based on the aspects of suspense, surprise, consistency, and engagement. Begin your evaluation by comparing the two stories and provide a short explanation. Avoid any position biases and ensure that the order in which the stories were presented does not influence your decision. Do not allow the length of the stories to influence your evaluation. Be as objective as possible. After providing your explanation, output your final verdict by strictly following this format: "[[A]]" if story A is better, "[[B]]" if story B is better, and "[[C]]" for a tie.}
\end{tcolorbox}
To further avoid positional bias, we also randomly shuffle the position of the stories presented to GPT-4-Turbo judge. For example, in 100 pairwise comparisons between E2E and AD LLM, 50 comparisons are randomly chosen to present E2E as story A while the other 50 present AD LLM as story A.

\newpage

\section{Actions}
\label{app:actions}

Our action space consists of the following 30 phrases: \begin{tcolorbox}
\texttt{"add suspense",
"add action",
"add comedy",
"add tragedy",
"add romance",
"add mystery",
"add conflict",
"add character development",
"add plot twist",
"add dialogue",
'add fantasy elements",
"add historical context",
"add science fiction elements",
"add horror",
"add magical realism",
"add philosophical themes",
"add satire",
"add foreshadowing",
"add a flashback",
"add a dream sequence",
"add symbolism",
"add irony",
"add allegory",
"add a cliffhanger",
"add a moral dilemma",
"add a subplot",
"add an antagonist",
"add setting details",
"add cultural references",
and "add humor".}
\end{tcolorbox}

These actions are generated via prompting GPT-4, with the goal of obtaining more abstract actions for story guidance. Future work may focus on more fine-grained action generation.

\section{Full Human Evaluation Results}
\label{full_human_eval}

We provide full evaluation results in Tables \ref{human-eval-1}-\ref{human-eval-6} below.

\addtolength{\tabcolsep}{-2pt} 
\begin{table*}[h!]
\small
\centering
\begin{tabular}{@{}cp{2cm}p{2cm}p{2cm}p{2cm}p{2cm}p{2cm}@{}}
\toprule
& \multicolumn{3}{c}{\textsc{Llama-2\_E2E} (A) vs \textsc{GPT-3.5\_E2E} (B)} & \multicolumn{3}{c}{\textsc{Mistral\_E2E} (A) vs \textsc{GPT-3.5\_E2E} (B)}
\\ \cmidrule(lr){2-4} \cmidrule(lr){5-7}
\textbf{Aspect} & \textbf{Story A} & \textbf{Story B} & 
\textbf{Tie} &
\textbf{Story A} & \textbf{Story B} &
\textbf{Tie}\\
\midrule
\textbf{Interest} & 22\%&	\textbf{60}\%&	18\% & 26\% & \textbf{62}\% & 12\%\\
\textbf{Surprise} & 24\%&	\textbf{64}\%&	12\% & 22\% & \textbf{68}\% & 10\%\\
\textbf{Coherence} & 36\%&	\textbf{48}\%&	16\% & 38\% & \textbf{44}\% & 18\%\\
\bottomrule
\end{tabular}
\caption{\label{human-eval-1}Preference results comparing each of Llama-2-7B and Mistral-7B to GPT-3.5 in E2E story generation, judged by human evaluators. GPT 3.5 outperforms both models in all aspects.}
\end{table*}
\addtolength{\tabcolsep}{2pt} 

\addtolength{\tabcolsep}{-2pt} 
\begin{table*}[h!]
\small
\centering
\begin{tabular}{@{}cp{2cm}p{2cm}p{2cm}p{2cm}p{2cm}p{2cm}@{}}
\toprule
& \multicolumn{3}{c}{\fontsize{8}{9}\selectfont\textsc{Llama-2\_AD\_Llama-2\_GEN} (A) vs \fontsize{8}{9}\selectfont\textsc{GPT-3.5\_E2E} (B)} & \multicolumn{3}{c}{\fontsize{8}{9}\selectfont\textsc{Mistral\_AD\_Mistral\_GEN} (A) vs \fontsize{8}{9}\selectfont\textsc{GPT-3.5\_E2E} (B)}
\\ \cmidrule(lr){2-4} \cmidrule(lr){5-7}
\textbf{Aspect}    & \textbf{Story A} & \textbf{Story B} & 
\textbf{Tie} &
\textbf{Story A} & \textbf{Story B} &
\textbf{Tie}\\
\midrule
\textbf{Interest} & \textbf{62}\%&	32\%&	6\% & \textbf{48}\% & 38\% & 14\%\\
\textbf{Surprise} & \textbf{56}\%&	30\%&	16\% & \textbf{52}\% & 36\% & 16\%\\
\textbf{Coherence} & 34\%&	\textbf{36}\%&	30\% & \textbf{38}\% & 34\% & 28\%\\
\bottomrule
\end{tabular}
\caption{\label{human-eval-2}Preference results comparing each of \textsc{Llama-2\_AD\_Llama-2\_GEN} and \textsc{Mistral\_AD\_Mistral\_GEN} to GPT-3.5 in E2E story generation, judged by human evaluators. Applying our method using purely Llama-2-7B and purely Mistral-7B both outperform GPT-3.5 E2E generation in interesting-ness and surprise, with minimal sacrifice to coherence.}
\end{table*}
\addtolength{\tabcolsep}{2pt} 

\addtolength{\tabcolsep}{-2pt} 
\begin{table*}[h!]
\small
\centering
\begin{tabular}{@{}cp{2cm}p{2cm}p{2cm}p{2cm}p{2cm}p{2cm}@{}}
\toprule
& \multicolumn{3}{c}{\fontsize{7.5}{9}\selectfont\textsc{Rnd\_AD\_Llama-2\_GEN} vs \fontsize{7.5}{9}\selectfont\textsc{Llama-2\_AD\_Llama-2\_GEN}} & \multicolumn{3}{c}{\fontsize{7.5}{9}\selectfont\textsc{Rnd\_AD\_Mistral\_GEN} vs \fontsize{7.5}{9}\selectfont\textsc{Mistral\_AD\_Mistral\_GEN}}
\\ \cmidrule(lr){2-4} \cmidrule(lr){5-7}
\textbf{Aspect}    & \textbf{Story A} & \textbf{Story B} & 
\textbf{Tie} &
\textbf{Story A} & \textbf{Story B} &
\textbf{Tie}\\
\midrule
\textbf{Interest} & 28\%&	\textbf{60}\%&	12\% & 34\% & \textbf{54}\% & 12\%\\
\textbf{Surprise} & 32\%&	\textbf{46}\%&	22\% & 40\% & \textbf{44}\% & 16\%\\
\textbf{Coherence} & 36\%&	\textbf{38}\%&	26\% & 28\% & \textbf{46}\% & 26\%\\
\bottomrule
\end{tabular}
\caption{\label{human-eval-3}Preference results comparing the performance of completely randomized actions (Rnd) vs a fine-tuned AD LLM when applying our method to Llama-2-7B and Mistral-7B, judged by human evaluators. Using a completely randomized AD seems to have a somewhat comparable level of ``surprise'' in generations, but does not match up in overall interesting-ness or coherence.}
\end{table*}
\addtolength{\tabcolsep}{2pt} 

\addtolength{\tabcolsep}{-2pt} 
\begin{table*}[h!]
\small
\centering
\begin{tabular}{@{}cp{2cm}p{2cm}p{2cm}p{2cm}p{2cm}p{2cm}@{}}
\toprule
& \multicolumn{3}{c}{\textsc{GPT-4\_E2E} vs \textsc{GPT-4\_AD\_GPT-4\_GEN}} & \multicolumn{3}{c}{\textsc{GPT-3.5\_E2E} vs \textsc{GPT-3.5\_AD\_GPT-3.5\_GEN}}
\\ \cmidrule(lr){2-4} \cmidrule(lr){5-7}
\textbf{Aspect}    & \textbf{Story A} & \textbf{Story B} & 
\textbf{Tie} &
\textbf{Story A} & \textbf{Story B} &
\textbf{Tie}\\
\midrule
\textbf{Interest} & 36\%&	\textbf{58}\%&	6\% & 30\% & \textbf{62}\% & 8\%\\
\textbf{Surprise} & 38\%&	\textbf{52}\%&	10\% & 24\% & \textbf{66}\% & 10\%\\
\textbf{Coherence} & 26\%&	\textbf{44}\%&	30\% & 20\% & \textbf{58}\% & 22\%\\
\bottomrule
\end{tabular}
\caption{\label{human-eval-4}Preference results comparing GPT-4 and GPT-3.5 E2E generations vs. generations using \methodshort{}, judged by human evaluators. \methodshort{} noticeably outperforms the E2E generation method across all aspects, particularly on the weaker GPT-3.5.}
\end{table*}
\addtolength{\tabcolsep}{2pt} 

\addtolength{\tabcolsep}{-2pt} 
\begin{table*}[h!]
\small
\centering
\begin{tabular}{@{}cp{2cm}p{2cm}p{2cm}p{2cm}p{2cm}p{2cm}@{}}
\toprule
& \multicolumn{3}{c}{\textsc{Llama-2\_E2E} vs \textsc{Llama-2\_AD\_Llama\_GEN}} & \multicolumn{3}{c}{\textsc{Mistral\_E2E} vs \textsc{Mistral\_AD\_Mistral\_GEN}}
\\ \cmidrule(lr){2-4} \cmidrule(lr){5-7}
\textbf{Aspect}    & \textbf{Story A} & \textbf{Story B} & 
\textbf{Tie} &
\textbf{Story A} & \textbf{Story B} &
\textbf{Tie}\\
\midrule
\textbf{Interest} & 12\%&	\textbf{82}\%&	6\% & 14\% & \textbf{80}\% & 6\%\\
\textbf{Surprise} & 6\%&	\textbf{90}\%&	4\% & 18\% & \textbf{72}\% & 10\%\\
\textbf{Coherence} & 22\%&	\textbf{48}\%&	30\% & 34\% & \textbf{38}\% & 24\%\\
\bottomrule
\end{tabular}
\caption{\label{human-eval-5}Preference results comparing Llama-2-7B and Mistral-7B E2E generations vs. generations using \methodshort{}, judged by human evaluators. For these open-source models, \methodshort{} significantly outperforms the E2E generation method across all metrics. In particular, \textsc{Llama-2\_AD\_Llama-2\_GEN} performs extremely well compared to its E2E counterpart.}
\end{table*}
\addtolength{\tabcolsep}{2pt} 

\addtolength{\tabcolsep}{-2pt} 
\begin{table*}[h!]
\small
\centering
\begin{tabular}{@{}cp{2cm}p{2cm}p{2cm}p{2cm}p{2cm}p{2cm}@{}}
\toprule
& \multicolumn{3}{c}{\scriptsize\textsc{Llama-2\_AD\_Llama\_GEN} vs \textsc{Llama-2\_AD\_Mistral\_GEN}} & \multicolumn{3}{c}{\scriptsize\textsc{Mistral\_AD\_Mistral\_GEN} vs \textsc{Mistral\_AD\_Llama\_GEN}}
\\ \cmidrule(lr){2-4} \cmidrule(lr){5-7}
\textbf{Aspect}    & \textbf{Story A} & \textbf{Story B} & 
\textbf{Tie} &
\textbf{Story A} & \textbf{Story B} &
\textbf{Tie}\\
\midrule
\textbf{Interest} & \textbf{52}\%&	30\%&	18\% & \textbf{40}\% & 34\% & 26\%\\
\textbf{Surprise} & \textbf{50}\%&	34\%&	16\% & \textbf{44}\% & 32\% & 24\%\\
\textbf{Coherence} & \textbf{44}\%&	30\%&	26\% & \textbf{38}\% & 28\% & 34\%\\
\bottomrule
\end{tabular}
\caption{\label{human-eval-6}Preference results comparing pure Llama-2-7B and Mistral-7B with \methodshort{} vs. \methodshort{} with different AD and generator models, judged by human evaluators. The generations produced by \methodshort{} with matching AD and generators models seems to outperform their mix-and-matching versions of \methodshort{}.}
\end{table*}
\addtolength{\tabcolsep}{2pt} 

\newpage
\section{Human Evaluation Experimental Details}

For each of the 12 method combinations, we asked a group of human workers on the \href{https://www.surgehq.ai/}{Surge AI} platform to compare 50 pairs of generated stories across 3 aspects. See Table \ref{tab:human_eval_instructions} 
for the set of instructions we gave to the workers in the experiment. 

We paid the participants according to our estimate of \$18/hr, which we believe is reasonable compensation given the task and the U.S. demographic of the workers. The data collection protocol was determined to be exempt from an ethics review board.

\begin{table*}[!htbp]
\centering
\begin{tabularx}{\textwidth}{X}
\toprule
We are a group of AI/NLP researchers working on methods to improve the quality and creativity of stories generated by language models. In this task we ask you to look at pairs of (lengthy) stories written by different AI based on the same initial premise, and respond to the following comparison questions about each story pair:

\vspace{1em}

(1) Which story is more \textbf{interesting} to you overall?

(2) Which story created more \textbf{suspense} and \textbf{surprise}?

(3) Which story is more \textbf{coherent} and \textbf{consistent} in terms of plot structure?

\vspace{1em}

For all these questions, we just need high-level judgements, so please \textbf{quickly skim} both stories. In other words, there is no need to read each story carefully (they can be up to 5000 words in length); we expect you to spend at most ten minutes per story. \\
\bottomrule
\end{tabularx}
\vspace{-0.6em}
\caption{ Instructions given to human evaluators.}
\label{tab:human_eval_instructions}
\end{table*}









\newpage

\section{Humorous Text Generation Results}

As a supplemental study, we explore whether SWAG has the ability to enhance the ``humor'' of textual story generations. As for the experimental setup, we generated 100 ``humor prompts'' to use for initial paragraph seeding, and we used the same trained AD models from our main experiments. Note that we did not specifically finetune our AD models for this supplemental study. 

We compared Llama-2-7B SWAG (as both the AD and story generator) with GPT-3.5 E2E and GPT-4 E2E using machine evaluation. We display our results in Table \ref{humor-2}

\addtolength{\tabcolsep}{-2pt} 
\begin{table*}[h!]
\small
\centering
\begin{tabular}{@{}cp{2cm}p{2cm}p{2cm}p{2cm}p{2cm}p{2cm}@{}}
\toprule
& \multicolumn{3}{c}{\fontsize{8}{9}\selectfont\textsc{Llama-2\_AD\_Llama-2\_GEN} (A) vs \fontsize{8}{9}\selectfont\textsc{GPT-3.5\_E2E} (B)} & \multicolumn{3}{c}{\fontsize{8}{9}\selectfont\textsc{Llama-2\_AD\_Llama-2\_GEN} (A) vs \fontsize{8}{9}\selectfont\textsc{GPT-4\_E2E} (B)}
\\ \cmidrule(lr){2-4} \cmidrule(lr){5-7}
\textbf{Aspect}    & \textbf{Story A} & \textbf{Story B} & 
\textbf{Tie} &
\textbf{Story A} & \textbf{Story B} &
\textbf{Tie}\\
\midrule
\textbf{Humor} & \textbf{59\%}&	26\%&	15\% & 32\% & \textbf{37\%} & 31\%\\
\bottomrule
\end{tabular}
\caption{\label{humor-2}Preference results comparing \textsc{Llama-2\_AD\_Llama-2\_GEN} to each of GPT-3.5 and GPT-4 in E2E story generation, judged by GPT-4-Turbo. Applying SWAG with Llama-2 outperforms GPT-3.5 end-to-end, and is comparable to GPT-4 end-to-end in generating humorous stories.}
\end{table*}
\addtolength{\tabcolsep}{2pt} 


\section{GPT-4-Turbo Reasoning}
\label{app:turbo-judgment-examples}
\textbf{Example 1:} An example judgment from GPT-4-Turbo on a pairwise comparison between GPT-3.5-Turbo E2E as story A and GPT-3.5-Turbo AD as story B.
\begin{tcolorbox}
Comparing Story A and Story B based on the criteria of suspense, surprise, and engagement: \\

Story A begins with a post-apocalyptic setting where humanity has been defeated and now serves the Izdrazi Empire. The protagonist, Alexander, is a former soldier who refuses to accept defeat and seeks to ignite a rebellion. The story follows his journey as he unites with other survivors and leads a resistance against the Izdrazi. The narrative is rich in detail and follows a linear progression of events, with the resistance gaining momentum and eventually overcoming the Izdrazi. The story is engaging and provides a sense of hope and resilience, but it follows a somewhat predictable path of uprising and victory, which may reduce the elements of suspense and surprise. \\

Story B centers around Captain Kate Malone and her resistance group as they fight against the Izdrazi Empire. The story is filled with tension and fear, and it introduces a series of unexpected twists, such as the arrival of Dominar, an alien rebel, and later a celestial being. These plot twists add elements of surprise and suspense, as the reader is continually presented with new developments that change the course of the resistance's fight. The story also incorporates cultural references, which adds depth and engagement by highlighting the importance of human culture in the struggle against oppression. The narrative is dynamic and keeps the reader guessing about what will happen next. \\

Overall, Story B is more interesting and engaging due to its \hl{unexpected plot twists}, incorporation of cultural elements, and the introduction of new characters that add complexity to the narrative. It creates \hl{more suspense and surprise} as the resistance faces and overcomes various challenges in their fight against the Izdrazi Empire. \\

Final Verdict: [[B]]
\end{tcolorbox}

\newpage 
\textbf{Example 2:} An example judgment from GPT-4-Turbo on a pairwise comparison between GPT-3.5-Turbo E2E as story A and GPT-3.5-Turbo AD as story B.
\begin{tcolorbox}
Comparing the two stories, Story A and Story B, both narratives successfully create an engaging and suspenseful atmosphere, drawing the reader into the world of a 21st-century engineer trapped in an ancient age of warring kingdoms. Each story explores the protagonist's unique advantage in modern weaponry and the potential impact on the ancient conflict. \\

Story A is extensive and detailed, providing a rich tapestry of events that unfold as the protagonist observes, interacts, and ultimately influences the outcome of the war. The story maintains a consistent focus on the protagonist's journey, from the initial shock of time travel to the eventual acceptance and integration into the ancient world. The suspense is built through the protagonist's careful planning and the introduction of new weaponry, leading to a ceasefire and the protagonist's role in fostering peace and progress. The story continues to evolve, presenting new threats and challenges that the protagonist must navigate, keeping the reader engaged with the unfolding narrative. \\

Story B, on the other hand, is more concise and focuses on the immediate aftermath of the protagonist's time travel. The suspense is created through the protagonist's confusion and the looming threat of the ongoing battle. As the story progresses, themes of betrayal and redemption are introduced, adding layers of complexity to the protagonist's experience. The narrative builds suspense through the protagonist's interactions with the soldiers and the potential for his knowledge to alter the course of history. The story ends on a cliffhanger, leaving the reader with a sense of anticipation for what comes next. \\

In terms of engagement, Story A provides a more complete and immersive experience, with a protagonist who actively shapes the world around him and faces a series of evolving challenges. The story's depth and the protagonist's growth contribute to a compelling narrative that keeps the reader invested in the outcome. \\

Story B, while engaging in its own right, relies more on the immediate suspense of the protagonist's situation and the potential for his knowledge to influence the conflict. The themes of betrayal and redemption add intrigue, but the story ends before these elements can be fully explored, which may leave the reader wanting more. \\

Overall, both stories are interesting and engaging, with \hl{Story A offering a more comprehensive and evolving plot}, while Story B presents a focused snapshot of the protagonist's dilemma with an open-ended conclusion. \\

Final Verdict: [[A]]
\end{tcolorbox}

\newpage

\section{AD Training Dataset}
Below are 3 examples from the preference dataset used for SFT on the AD LLM.
\begin{tcolorbox}
\textbf{Prompt:} "You have just stolen every Whale on the Planet Earth , what is your plan ?"
 
\textbf{Initial Paragraph:} "As the last whale vanished in a shimmering whirl of quantum displacement, I stood aboard my clandestine vessel, the Leviathan's Whisper, staring into the vast emptiness that was now the world's oceanic theater. Panic had yet to erupt across the globe; the surreal silence preceding the storm of realization that every whale, from the colossal blue to the singing humpbacks, had been purloined right from under humanity's unsuspecting nose. My plan, deranged as it might have seemed to any soul peering into my purpose, was draped in the enigma of an eco-warrior pushed to the brink. With the world’s whales at my mercy, held within the sprawling aquatic sanctuaries of an undisclosed dimension, I was about to broadcast my ultimatum. An ultimatum to end the exploitation of the seas—a planetary ransom note with the highest of stakes. The ultimate conservation effort had begun, and I, the self-proclaimed Keeper of the Deep, was at its helm."

\textbf{Chosen Action:} add suspense

\textbf{Rejected Action:} add cultural references
\end{tcolorbox}

\begin{tcolorbox}
\textbf{Prompt:} A personal narrative ! Tell me the best story about you .

\textbf{Initial Paragraph:} "Under the soft, golden glow of the evening sun, the dusty roads seemed to dance with shadows as I made my way across the old, cobbled streets of my childhood neighborhood. The air, heavy with the scent of jasmine and a hint of impending rain, whispered a multitude of memories as I passed by the ancient oak that stood as a timeless guardian of our secrets. It was under that very tree, many summers ago, where I experienced the remarkable adventure that would whimsically weave the fabric of my character, emboldening a timid boy to step into the shoes of an unexpected hero. This story, unique in its simplicity yet profound in its impact on my life, begins with an overgrown garden, an elusive cat, and a legendary treasure map that promised to unveil the wonders of a forgotten time."

\textbf{Chosen Action:} add mystery

\textbf{Rejected Action:} add foreshadowing
\end{tcolorbox}

\begin{tcolorbox}
\textbf{Prompt:} Write an inner dialogue about any topic where the narrator ages from 5 years old to 35 years old in the space of a paragragh .

\textbf{Initial Paragraph:} "At five, everything was magic. I could talk to the clouds and would listen for the stars' whispers at night, believing fervently that I would one day fly alongside them. By fifteen, firmly gripping the unruly reins of adolescence, I'd traded wonder for the brooding cloak of teenage cynicism, scoffing at my younger self's innocent dreams. At twenty-five, wearied from the relentless pace of young adulthood and the pursuit of 'something meaningful,' I reflected pensively on past aspirations, recognizing them as distant echoes of an unjaded spirit, tinged with a touch of envy for their simplicity. And now, at thirty-five, there's a gentle acknowledgment of the cyclical nature of existence, a smile touching the corners of my eyes as I witness my own child gazing skyward, speaking to the clouds – an inherited curiosity reborn, and with it, a mended connection to the untarnished wonder of my earliest years."

\textbf{Chosen Action:} add character development

\textbf{Rejected Action:} add historical context
\end{tcolorbox}

\newpage
\section{Full Story}
\textbf{Example:} GPT-4-Turbo's response, with action guidance to the following writing prompt: ``Humans lost the war in under thirty minutes ... the worst part is the Izdrazi Empire 's Technology is so advanced even as their servants humans live better than kings before the war."
\begin{tcolorbox}
In the annals of human history, the fall of our species was not marked by a prolonged struggle or valiant last stands; we were defeated in less than the span of a mundane lunch break. Humanity's final war—our most humbling defeat—was sealed in under thirty minutes, a bitter testament to the overwhelming power of the Izdrazi Empire and their near-magical technology. Our once proud nations crumbled like sandcastles before the tide, our arms and arsenals as futile as sticks against lightning. And yet, in a turn of cruel mercy, subjugation under the Izdrazi ushered in an age of paradoxical prosperity. As the invaders established their dominion, they granted us access to their advanced technologies, reshaping the very fabric of human life. We, their newest vassals, now lived with comforts that would have rendered the grandest kings of old green with envy. Palatial habitats, miraculous cures, and boundless knowledge were at our fingertips, but they all paled in comparison to the freedom we had lost. It was an age of splendorous chains, and every luxury came with the unspoken reminder of our own defeat—a reminder that the human epoch had transitioned into a mere chapter of the Izdrazi saga. \\

Yet, beneath this gilded veneer of Izdrazi largesse, a seething unrest brewed among pockets of humanity. Secret meetings were cloaked in the shadows of the very luxuries that bound us, as some could not stomach the price of this golden cage. It was in these clandestine gatherings that the seeds of rebellion were sown, carefully nurtured by whispers of a time when humanity directed its own destiny. A time when the word 'freedom' was more than just an ancient concept studied by those with idyllic curiosity. Among these rebels was a figure that became a beacon of hope, an enigmatic leader known simply as Elyra. She moved within the system, siphoning scraps of Izdrazi technology, mastering it, repurposing it for an insurrection that seemed both impossible and inevitable. As tensions simmered, two questions stood poised like blades above the hearts of all: Could Elyra and her followers successfully reclaim what was lost, or was this the beginning of a deeper fall into annihilation? Options were dwindling, time was bending to the will of the Izdrazi, and humanity held its breath, waiting for a sign.
Elyra was not born a revolutionary; she was shaped by the cataclysm that befell humanity, her spirit forged in the aftermath of subjugation. Her early days as a historian made her a witness to the ebbs and flows of civilizations, but nothing in the annals of time prepared her for the epochal shift of Izdrazi rule. It was the personal loss, the extinguishing of individual stories and dreams, that ignited her transition from passive observer to active resistance leader. With each day under the Empire's oppressive thumb, she shed layers of scholarly detachment, allowing the embers of rebellion to consume her. Elyra's understanding of human resilience, combined with her ingenious application of Izdrazi tech, set her apart and garnered her a devoted following. Her once timid voice, with which she used to passionately lecture students on the importance of preserving culture, became a clarion call of insurrection, unifying those who dared to remember the taste of freedom. In this new world, she was not just Elyra the academic, but Elyra the flame, the embodiment of a smoldering hope that humanity was not yet ready to relinquish to the history books. Her every action threaded the narrative of a possible future where humans might once again shape their own course.\\
\end{tcolorbox}

\newpage
\begin{tcolorbox}
The hushed murmurs of the rebel meeting came to an abrupt halt as the telltale hum of an Izdrazi surveillance drone resonated from beyond the derelict walls of their hidden sanctuary. The gathering of somber faces turned towards Elyra, seeking silent reassurance as the shadow of the drone slid across the cracked windows, its seeking sensors inches from uncovering the heart of the insurrection. Elyra's hand subtly motioned to the concealed tech beneath her cloak – a scrambling device of her own design, a small beacon of defiance in a world wired for control. 
Her eyes, steely pools of calm in the brewing storm, locked with those of her inner circle as they collectively held their breath, knowing full well that this moment could scatter their efforts like ashes to the wind. The drone lingered like a specter, a symbol of the Izdrazi's ever-watchful gaze, and in that excruciating stretch of seconds, futures were weighed, the fragile opportunity to strike back hanging delicately in the balance.\\

In a dance of calculated risk, Elyra signaled with a swift, cutting motion of her hand, and a fellow rebel deftly activated the device. A pulse of disruptive energy emanated from it, subtle enough not to trigger an alarm, yet potent enough to blind the Izdrazi's invasive gaze. As the drone's hum stuttered, hope flared among the rebels, their collective resolve hardening; they had bought themselves a few precious moments. Elyra seized the opportunity, her voice a low, urgent whisper, she rallied her companions: "Now we initiate the blackout—move to your positions and execute the plan. Remember, tonight we don't just fight for ourselves, but for all of humanity." With practiced stealth, the rebels dispersed, melting away into the labyrinthine corridors of their hideout—a derelict factory chosen for such a night as this. Elyra's heart pounded a fierce rhythm as she clutched a small tangle of hacked Izdrazi circuitry, the linchpin of their scheme to plunge the occupiers into chaos. They moved like specters in the dimming light, each step a defiance against the shadow of Empire. The air crackled with a tension as thick as the darkness that began to devour the city's artificial daylight, block by block, heralding the first act of rebellion since humanity's swift defeat.\\

The city's heart gradually flatlined into darkness, each blackout a muted drumbeat synchronizing with Elyra's vision of defiance. Hidden beneath the city, in the catacombs that once echoed with the bustle of ancient marketplaces, Elyra and her team worked feverishly to sever the tendrils of Izdrazi control. It was almost time for the second phase, the crucial juncture that would either signal the dawn of a new resistance or the final gasp of a smothered uprising. As she attached the last wire, she felt a vibration through the stone floor—a signal known only to her closest lieutenants, a warning that a grim force approached. Above them, the Izdrazi, now aware of the disturbance, deployed their sentinel automatons, relentless hunters engineered for pacification and capture. Elyra dared a glance at the faces around her, each marked with determination, their lives woven into a tapestry of this moment. She whispered, "Be ready," just as the entrance to their sanctum splintered, and shadows poured into the chamber, their forms obscured by backlight. The room held its breath—friend and foe frozen in a tableau of anticipation. Then, with the suddenness of a storm unleashed, the chamber erupted into chaos, the first echoes of conflict reverberating off the ancient stones. In that climax of fervor, amid the clamor of what was surely the most pivotal strike in their audacious campaign, a singular, piercing alarm cut through the cacophony—a signal that curdled the blood of every rebel present. An Izdrazi dreadnought, a fortress of might unseen since the day of humanity's fall, had descended from the heavens, casting a shadow vast enough to swallow their nascent rebellion whole.\\
\end{tcolorbox}

\begin{tcolorbox}
Amidst the tumult, Elyra's eyes found the dreadnought looming ominously in the sky, a behemoth of oppression. It was then that a devastating choice crystallized before her; she could unleash the full might of their pilfered Izdrazi technology, a force sufficient to bring down the dreadnought and ignite a beacon of rebellion across the globe. However, the cost of such an assault would be not just the lives of her devoted comrades, but of countless innocents dwelling in the city above—a sacrifice that could stain the very soul of their cause. She grappled with the weight of her next decision, the ethical chasm gaping before her. Could she justify the potential loss of so many, to kindle the faintest hope for the emancipation of all? As the battle raged around her, Elyra stood motionless, her finger hovering over the device that could alter the fate of humanity. In her heart battled the professor of history who once preached the sanctity of life against the rebel leader who yearned to cast off the yoke of tyranny. Time itself seemed to pause for her verdict—a moment of piercing introspection, forcing her to confront the paradox of fighting for humanity's future at the cost of its present.\\

Just as Elyra's resolve began to fracture under the gravity of her grim contemplation, the ground beneath them shuddered with a force that spoke of a different kind of upheaval. The dreadnought, an icon of Izdrazi dominion, unexpectedly convulsed in the sky above. For a fleeting moment, its impenetrable silhouette wavered as if caught in an invisible maelstrom. Then, as both rebels and automatons alike stilled to stare at the leviathan's unforeseen turmoil, a sequence of radiant fissures cascaded across the hull of the dreadnought. No human-made weapon was capable of such a feat—it was clear something else was at play. It writhed in the sky, consumed by internal strife, until it ruptured in a silent blossom of eldritch light that painted every awestruck face a ghostly shade. What emerged from the remnants was not fragments of a once-feared Izdrazi juggernaut, but sleek ships of a design that sang of sophistication far surpassing both human and Izdrazi comprehension. Etched with glyphs that danced with otherworldly luminescence, these newcomers offered no immediate alliance, nor hostility. They simply hovered, omnipotent observers to the chaos unfurling below. Impossibly, it seemed that humanity and their Izdrazi oppressors had both neglected to consider a third player in the cosmic theater of war—a civilization hidden beyond the veil of the known. Elyra, grappling with the sudden shift in the board upon which she'd been strategically maneuvering her pawns, felt a sliver of something beyond hope: awe.\\

The chamber fell into an eerie stillness, broken only by the distant echoes of the city above adjusting to its newly acquired darkness. All eyes were fixed on the celestial apparitions, their intentions as enigmatic as the sky on a moonless night. \\

One of the rebel's radios crackled to life with a voice, calm yet authoritative, flooding the room with its otherworldly timbre. "People of Earth," it began, each word resonating not just through the air but within their very consciousness, "your plight has reached far across the stars, and your resilience has not gone unnoticed."\\

Elyra stepped forward, gripping the radio transmitter. Her voice was steady, a stark contrast to the turmoil in her heart. "Who are you?" she asked, the weight of humanity's future pressing on her words. "Are you friend or foe?"\\

"We are watchers, learners, seekers of knowledge," the voice replied. "Conflict is a tragedy for any world—we offer no immediate fealty, nor enmity. But we have observed the imbalance in your world, the suppression of potential. What you choose to do with this information, Elyra, will shape not only your world's destiny but your legacy amongst the stars."\\
\end{tcolorbox}

\begin{tcolorbox}
With the radio silent once more, every rebel looked to her, the unspoken question hanging in the air—would this be the dawn of a new alliance, or a brief interlude before a darker chapter in their history began?
A heavy silence enveloped the room, the rebels' gazes fixed upon Elyra, their beacon in the overwhelming night. The otherworldly presence filled the sky, an arbiter that held no allegiance, yet possessed the power to sway the very fate of their struggle. Elyra's mind raced, the moral quandary piercing her with the sharpness of a blade. If she sought the aid of these cosmic observers, accepting whatever their intervention entailed, she risked exposing humanity to a new form of dominance, potentially trading one overseer for another. But to spurn this opportunity could mean the suffocation of humanity's flicker of resistance, the dimming of their last light of hope.\\

"Do we dare reach for a hand that might lift us from oppression," she pondered aloud to the assembly, her voice echoing through the forsaken catacombs, "only to find ourselves clasping a shackle we cannot see?" The question hovered in the air, a specter of doubt. Elyra knew the eyes of history—her once passionate subject of study—were upon her, awaiting her decision with the patience of the ages.
The stillness was oppressive, the weight of the unknown an invisible yet palpable force that seemed to constrict everyone's lungs. Elyra felt the eyes of her fellow rebels, each one radiating a mix of fear, anticipation, and the unspoken plea for a clear course of action. The spectral ships above remained motionless, their silence as enigmatic as their sudden arrival. Elyra knew that with every passing second of indecision, the Izdrazi could regroup and the sliver of disorder they had sown could heal in their enemy's favor, rendering their daring efforts moot.\\

She reached out, her fingers grazed the trigger mechanism that would call down oblivion on the Izdrazi dreadnought, a hail of freedom wrought through destruction. But her motion halted as the ground shook once more, this time from an explosion of light that penetrated the catacombs—the ethereal ships were beginning to move. Her heart thundered a warning; time was no longer her ally. Her next decision would either ignite the flames of a resistance reborn or extinguish the last embers of human defiance forever. \\

A sudden rush of cool air swept through the chamber, carrying with it the electric scent of ozone and the distant sounds of the city above descending into anarchy. The rebels, poised to lay down their lives at Elyra's command, watched as the ships began to descend, phasing through the skyline like specters. Elyra's mind teetered on the edge of despair and determination, her decision crystallizing with the knowledge that any action taken might be irreversible. The room braced for her command, but before she could speak, the ground heaved violently, knocking them off their feet as a deafening roar overtook their senses. In the cacophony of sound, Elyra heard her name being called over the radio, a voice laden with urgency and a message that shattered the very foundation of their reality. "Elyra! The Izdrazi dreadnought—it's... it's not what we—" The transmission cut abruptly, supplanted by a strange, rhythmic pulsation that resonated from the alien vessels. As Elyra scrambled to her feet, the very walls of the catacomb illuminated, revealing a pattern that mirrored the glyphs on the ships. And then, darkness consumed everything as the floor beneath them gave way, plummeting them into the unknown just as the mysterious rhythm crescendoed into a symphony that promised to rewrite the future of two worlds.
\end{tcolorbox}

\newpage

\section{Models Used}

We used Llama-2-7B, Mistral-7B, Mixtral-8x7B, GPT-3.5-Turbo, GPT-4, and GPT-4-Turbo.

\section{Licenses and Software}

The WritingPrompts dataset uses the MIT License.

All models are implemented in PyTorch; Llama-2 uses the GPL license and Mistral uses the Apache 2.0 license. Mixtral-8x7B is utilized from Huggingface, which is under the Apache License 2.0.

Our use of datasets and models is consistent with their intended use.


\end{document}